\documentclass{article}

\usepackage[dvipsnames]{xcolor} 

\newcommand{\hiddencomment}[1]{}  

\definecolor{CMpurple}{rgb}{0.6,0.18,0.64}
\newcommand{\cm}[1]{} 

\newcommand{\secref}[1]{Section~\ref{#1}}
\newcommand{\tabref}[1]{Table~\ref{#1}}
\newcommand{\figref}[1]{Figure~\ref{#1}}
\newcommand{\appref}[1]{Appendix~\ref{#1}}

\usepackage[final]{ap_2020}

\usepackage[utf8]{inputenc} 
\usepackage[T1]{fontenc}    
\usepackage{hyperref}       
\usepackage{url}            
\usepackage{booktabs}       
\usepackage{amsfonts}       
\usepackage{nicefrac}       
\usepackage{microtype}      
\usepackage{wrapfig}
\usepackage{graphicx}
\usepackage{float}
\usepackage{subcaption} 
\usepackage{amsmath}
\usepackage{longtable}
\usepackage{tikz}
\usepackage{ctable}         
\usepackage{afterpage}      
\usepackage{setspace}

\usepackage{color, colortbl}

\definecolor{LightCyan}{rgb}{0.88,1,1}
\definecolor{aquamarine}{rgb}{0.5, 1.0, 0.83}
\definecolor{babyblue}{rgb}{0.54, 0.81, 0.94}
\definecolor{babypink}{rgb}{0.96, 0.76, 0.76}
\definecolor{corn}{rgb}{0.98, 0.93, 0.36}
\definecolor{vanilla}{rgb}{0.95, 0.9, 0.67}
\definecolor{timberwolf}{rgb}{0.86, 0.84, 0.82}

\newcommand{\customparagraph}[1]{\textbf{#1} \hspace{4pt}}

\title{Neural Generation Meets Real People: Towards Emotionally Engaging Mixed-Initiative Conversations}

\author{
  Ashwin Paranjape\thanks{equal contribution}, 
  Abigail See,\footnotemark[1] \hspace{0.1em} 
  Kathleen Kenealy, Haojun Li, Amelia Hardy, Peng Qi, \\
  \textbf{Kaushik Ram Sadagopan, Nguyet Minh Phu, Dilara Soylu, Christopher D. Manning} \\ \\
  Stanford NLP \\ \\
  \texttt{\{ashwinpp,abisee,kkenealy,haojun,ahardy,pengqi,} \\
  \texttt{kaushik7,minhphu,soylu,manning\}@stanford.edu} \\
}

\begin{document}

\maketitle

\begin{abstract}
We present \textbf{Chirpy Cardinal}, an open-domain dialogue agent, as a research platform for the 2019 Alexa Prize competition.
Building an open-domain socialbot that talks to real people is challenging -- such a system must meet multiple user expectations such as broad world knowledge, conversational style, and emotional connection. 
Our socialbot engages users on their terms -- prioritizing their interests, feelings and autonomy. 
As a result, our socialbot provides a responsive, personalized user experience, capable of talking knowledgeably about a wide variety of topics, as well as chatting empathetically about ordinary life.
Neural generation plays a key role in achieving these goals, providing the backbone for our conversational and emotional tone.
At the end of the competition, Chirpy Cardinal progressed to the finals with an average rating of 3.6/5.0, a median conversation duration of 2 minutes 16 seconds, and a 90$^{\text{th}}$ percentile duration of over 12 minutes. 
\end{abstract}

\section{Introduction}
\label{sec:intro}

This paper describes our socialbot for open-domain conversation, \textbf{Chirpy Cardinal}, built as a research platform during the 2019 Alexa Prize competition. 
During the competition, US-based Amazon Alexa users could give an invocation phrase (such as \textit{let's chat}) to be connected to one of the competing socialbots (chosen randomly).
After receiving a minimal orientation phrase at the beginning of the conversation, the user talks to the socialbot (in English) until they decide to end the conversation -- at which point, they are invited to provide a rating and comment.

To provide a convincing user experience, an open-domain conversational agent must excel at language understanding, language generation, emotional engagement, memory, world knowledge and conversational planning, among other desirable characteristics -- an ambitious goal!
Prior work within and outside the Alexa Prize competition has taken the successful strategy of pushing progress along individual skills, and forming an ensemble of sub-systems, each excelling at a singular characteristic while ignoring others. 
For instance, supporting user initiative in open-domain conversations is extremely challenging, as it requires understanding the countless ways a user can take initiative, and the ability to respond to each of them with specificity. 
Faced with this difficulty, when it comes to in-depth conversations, many previous dialogue systems rely primarily on bot-initiative, driving users along carefully scripted paths. 
On the other hand, systems attempting higher user-initiative via non-scripted paths are likely to lead towards shallower conversations. 
Thus there is a lot of room for innovation and research in trying to simultaneously achieve two or more complementary characteristics; this is a recurring theme throughout this work. 
Our goal in building this socialbot was to offer a natural-sounding and emotionally engaging dialogue agent that can talk knowledgeably about a wide variety of topics, while also letting the user take as much initiative as possible. 

\textbf{Initiative} -- the ability to drive the direction of the conversation -- has been studied extensively in the context of task-oriented dialogue. 
\textbf{Mixed initiative} \citep{horvitz1999principles}, in which the user and the bot share initiative, is an important quality of a successful dialogue system, as it provides the user a sense of agency without making them entirely responsible for suggesting new topics and directions.
In order to improve on mixed initiative while still providing an acceptable conversational depth, we designed our initial system to rely heavily on system initiative, but at the same time explored several avenues to increase user initiative in a controlled fashion. 
To support mixed initiative, our system has a global navigational intent classifier (\secref{sec:navigational_intent}) and entity tracker (\secref{sec:entity_tracker}), allowing it to track high level topic changes from both the user and the bot.
Further, our response priority system (\secref{sec:response_priority}) allows individual Response Generators (RGs) to interject when the user initiates a change of topic.

\textbf{High-coverage world knowledge} is an important component of open-domain conversation -- our bot must be able to talk about the diverse range of entities and topics that interest users, particularly if we wish to respect user initiative.
We use the Alexa Knowledge Graph, The Washington Post, Reddit and Twitter as sources of up-to-date knowledge in particular domains, while ensuring high coverage by using Wikipedia and Wikidata entities as the foundation of our entity-based conversations (Sections \ref{sec:entity_linker}, \ref{sec:entity_tracker} and \ref{sec:analysis_entitytracker}).
However, world knowledge must be delivered in a \textbf{conversational style} -- this is a characteristic that distinguishes a socialbot from a virtual assistant.
To achieve this, we finetuned a neural generative model on the TopicalChat dataset \citep{gopalakrishnan2019topical} to obtain a conversational paraphrasing model that adapts external text into a conversational style (\secref{sec:wiki}).

A socialbot cannot focus solely on external entities -- to be truly \textit{social}, it must be able to discuss \textbf{personal experiences and emotions}. While ELIZA-like systems \citep{eliza} attempt this via templated repetition of user phrases, they lack the naturalness and depth of real human conversations. 
Our Neural Chat module (\secref{sec:neural_chat}) invites the user to share their everyday experiences and current emotions, and uses a neural generative model to respond empathetically.
With it, we attempt to have a deep, sustained and emotionally engaging conversation about a user's lives.
In addition, our Opinion module (\secref{sec:opinion}) allows the user to express their feelings by expressing their likes and dislikes. 
To foster a reciprocal atmosphere, our bot also shares its own distinct feelings, experiences and opinions.

Lastly, we note that the advent of large-scale pretrained \textbf{neural generative models} has substantially impacted what is possible in open-domain socialbots. 
While in the last Alexa Prize competition, none of the top three socialbots used neural generation \citep{chen2018gunrock, pichi2018alquist, curry2018alana}, we found current GPT-2 models \citep{radford2019language} to be a key tool to support our design goals.
Neural generation enables natural phrasing and emotional engagement, as well as more flexible responsiveness (e.g., when used as a fallback in Section \ref{sec:neural_fallback}), supporting higher user initiative. 
A limitation of neural generation methods for dialogue is deterioration in quality and consistency over a long conversation, which can be potentially overcome with symbolic constraints.
We explore ways to bring the best of both worlds -- long term consistency and short term fluidity -- together. 

Despite being a first-time entrant, at the end of the competition our system achieved a rating of 3.6/5.0, which is within 0.1 of the highest-ranked systems, and is capable of detailed, sustained conversations with interested users (with a 90$^{\text{th}}$ percentile conversation duration of 12 minutes 55 seconds).
Qualitatively, during in-person interactions with users, we observed that many innovations such as in-depth discussions of everyday life, conversational styling of informational content, and opinionated exchanges were received with expressions of pleasant surprise -- indicating our steps were in the right direction. 
In \secref{sec:analysis}, we re-examine the goals we set out to achieve, and empirically analyze our bot's successes and failures.
In \secref{sec:discussion}, we talk about the challenges we faced, the trade-offs we made, our conclusions and avenues for future work. 

\begin{figure*}[t]
\centering
\includegraphics[width=\textwidth]{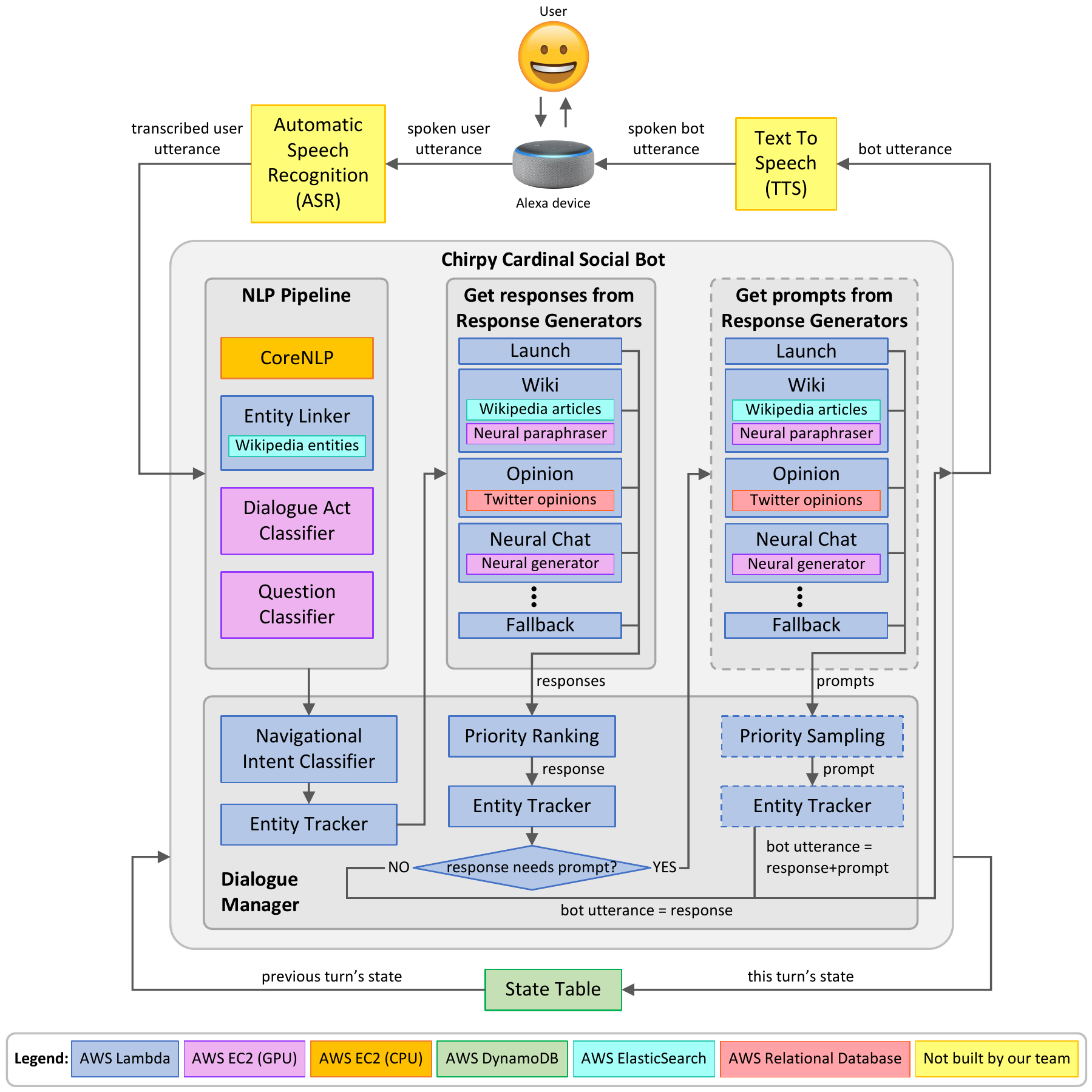}
\caption{Overall system design.}
\label{fig:overview_diagram}
\end{figure*}

\section{System Overview}
\label{sec:system_overview}

Our overall system design is shown in \figref{fig:overview_diagram}. 
Our system is built on top of the CoBot framework \citep{khatri2018advancing}.
On each turn, the user's spoken utterance is transcribed by Alexa's Automatic Speech Recognition (ASR) service. 
The transcribed utterance (which is lowercase, no punctuation) is sent to our AWS Lambda function, which handles the core logic of our bot. 
AWS Lambda is a serverless computing platform, which means that our function is stateless.
To preserve information between turns, we store our bot's overall state in an external State Table (see \figref{fig:overview_diagram}), hosted on AWS DynamoDB\@.
At the start of the turn, the previous turn's state is fetched from the table.

We then run the \textbf{NLP Pipeline} (see \secref{sec:nlp_pipeline}) -- a collection of modules that produce annotations based on the user's utterance and the current state.
Modules requiring greater computational resources are hosted on remote EC2 instances, while less-demanding modules are hosted within the Lambda function.
The NLP Pipeline is organized as a directed acyclic graph (DAG), allowing modules to use other modules' annotations as inputs. 
To minimize latency, modules are run in parallel where possible, with each module starting as soon as its inputs are ready.

Next, we analyze the user's utterance to determine whether the user wants to talk about any particular entity (see \textbf{Navigational Intent}, \secref{sec:navigational_intent}), and update the current entity under discussion if appropriate (see \textbf{Entity Tracker}, \secref{sec:entity_tracker}).

We then run our collection of \textbf{Response Generators} (RGs), modules designed to handle particular conversational duties, in parallel (see \secref{sec:rgs}). 
Each RG either produces a \textbf{response}, or no response (None).
If an RG produces a response, it also supplies a \textbf{response priority} (see \secref{sec:response_priority}), indicates whether the response needs a \textbf{prompt} added from another response generator (see \secref{sec:response_and_prompt}), and specifies what the current entity under discussion should be, if the response is chosen.
The \textbf{Priority Ranking} module chooses the response with the highest priority, and the Entity Tracker updates the current entity under discussion accordingly. 
If the chosen response \textit{does not} need a prompt, it forms the entire bot utterance.

If the chosen response \textit{does} need a prompt, we run our collection of RGs a second time.
Each RG either produces a prompt or no prompt (None). 
If an RG produces a prompt, it also supplies a \textbf{prompt priority} (see \secref{sec:prompt_priority}) and a current entity, as before.
The \textbf{Priority Sampling} module chooses the prompt by sampling from the supplied prompts, with the probability distribution depending on both the priorities of the prompts and the RGs that produced them.
The Entity Tracker updates the current entity again, and the bot's utterance is then formed by appending the prompt to the response.

At the end of the turn, the bot's overall state contains the user's utterance, the conversational history, the NLP Pipeline annotations for the user's utterance, and a state for each individual Response Generator.\footnote{Each individual RG state contains information required by that RG -- for example, it might contain the current node (aka treelet, see \secref{sec:treelets}) in the RG's dialogue graph, or a list of the utterances and/or entities that have been discussed, to avoid repetition.
Each RG maintains its own individual state.}
We write the new state to the State Table, and send the bot utterance to Alexa's Text To Speech (TTS) service, which delivers the spoken bot utterance to the user.

\afterpage{

\footnotesize

\setstretch{0.5}

\begin{longtable}[h1]{@{} p{0.02cm} p{6.5cm}  p{1.4cm}  p{5.7cm} @{}}
 \textbf{\#} & \textbf{Utterance} & \textbf{Entity} & \textbf{Commentary} \\
 \specialrule{.10em}{.1em}{.1em} 
 
    1 & \textbf{User:} let's chat & None & \\[0.8ex] 
    
    & \textbf{Bot:} Hi, this is an Alexa Prize Socialbot. I'd love to get to know you a bit better before we chat! Is it all right if I ask for your name? \colorbox{Apricot}{[ Launch ]} & None & \colorbox{Apricot}{Launch} gives \textit{FORCE\_START} response to begin the conversation. \\
    
    \specialrule{.01em}{.1em}{.1em} 

    2 & \textbf{User:} my name is chris & None & \\[0.8ex] 
    
    & \textbf{Bot:} Well it's nice to meet you, Chris! I'm excited to chat with you today. \colorbox{Apricot}{[ Launch ]} I hope your afternoon is going well. What are your plans for the rest of today? \colorbox{Aquamarine}{[ Neural Chat ]} & None & \colorbox{Apricot}{Launch} gives a \textit{STRONG\_CONTINUE} response that needs a prompt, which is supplied by \colorbox{Aquamarine}{Neural Chat} with \textit{FORCE\_START} priority (scripted transition). \\
    
    \specialrule{.01em}{.1em}{.1em} 
    
    3 & \textbf{User:} hang out with my friends & None & \\[0.8ex] 
    
    & \textbf{Bot:} That sounds great, I love spending time with friends. What will you do together? \colorbox{Aquamarine}{[ Neural Chat ] } & None & \colorbox{Aquamarine}{Neural Chat} generates a GPT-2 response with \textit{STRONG\_CONTINUE} priority. \\
    
    \specialrule{.01em}{.1em}{.1em} 
    
    4 & \textbf{User:} maybe watch a movie & None & \\ [0.8ex]  
    
    & \textbf{Bot:} Me too! I love watching movies; I get to learn so much about what the world is like outside of the cloud! Have you seen any movies recently? \colorbox{GreenYellow}{[ Movies ]} & Film & \colorbox{GreenYellow}{Movies} detects the `movie' keyword in the user's utterance and gives a \textit{FORCE\_START} response (overriding \colorbox{Aquamarine}{Neural Chat}), setting current entity to \textit{Film}. \\
    
    \specialrule{.01em}{.1em}{.1em} 
    
    5 & \textbf{User:} i saw the matrix & The Matrix & The \colorbox{Yellow}{Entity Linker} and \colorbox{Yellow}{Entity Tracker} identify \textit{The Matrix} as the current entity. \\[0.8ex] 
    
    & \textbf{Bot:} Nice! Did you like The Matrix? \colorbox{GreenYellow}{[ Movies ]} & The Matrix &  \\ 
    
    \specialrule{.01em}{.1em}{.1em} 
    
    6 & \textbf{User:} i loved it neo is amazing & The Matrix & \\[0.8ex] 
    
    & \textbf{Bot:} Oooh, yeah, I agree. Hey, isn't Keanu Reeves in that movie? What do you think about Keanu Reeves? \colorbox{GreenYellow}{[ Movies ]} & Keanu Reeves &  \colorbox{GreenYellow}{Movies} sets current entity to \textit{Keanu Reeves}. \\
    
    \specialrule{.01em}{.1em}{.1em} 
    
    7 & \textbf{User:} i want to talk about something else & None & \colorbox{Yellow}{Navigational Intent} detects \textit{negative intent}, so \colorbox{Yellow}{Entity Tracker} rejects the current entity. \\ [0.8ex]   

    & \textbf{Bot:} OK, no problem. \colorbox{GreenYellow}{[ Movies ]} There's actually something else I wanted to ask you about. What's your favorite animal? \colorbox{Dandelion}{[ Categories ]} & Animal & \colorbox{GreenYellow}{Movies} acknowledges and asks for a prompt. \colorbox{Dandelion}{Categories} gives a \textit{GENERIC} prompt. \\ [1ex] 
    
    \specialrule{.01em}{.1em}{.1em} 
    
    8 & \textbf{User:} i love cats & Cat & \colorbox{Yellow}{Entity Linker} and \colorbox{Yellow}{Entity Tracker} identify \textit{Cat} as current entity.  \\[0.4ex] 

    & \textbf{Bot:} Good to hear you like cats. I have to be honest though, I’m not a big fan of cats. I feel like cats don’t respect my personal space, but I would love to hear why you like cats? \colorbox{CarnationPink}{[ Opinion ]} & Cat & \colorbox{CarnationPink}{Opinion} gives a \textit{CAN\_START} response about cats. \\ 
    
    \specialrule{.01em}{.1em}{.1em} 
    
    9 & \textbf{User:} hmm i love cats because they are fluffy & Cat & \\ [0.4ex] 
    
    & \textbf{Bot:} That make sense. Now that I think about it, one good reason to like cats is that they purr and I definitely need that kind of positive feedback. \colorbox{CarnationPink}{[ Opinion ]} Wanna know something interesting about cat? \colorbox{SkyBlue}{[ Wiki ]} & Cat & \colorbox{CarnationPink}{Opinion} finishes talking. \colorbox{SkyBlue}{Wiki} gives a \textit{CURRENT\_TOPIC} prompt about \textit{Cat}. \\
    
    \specialrule{.01em}{.1em}{.1em} 
    
    10 & \textbf{User:} you are not very smart & Cat & \\ [0.4ex] 
    
    & \textbf{Bot:} I know you feel frustrated. I'm always trying to get better. \colorbox{Salmon}{[ Offensive User ]} I've been listening to some new music today and I wanted to chat about instruments.  If you were a musical instrument which one would you be? \colorbox{BurntOrange}{[ Music ]} & Musical instrument & \colorbox{Salmon}{Offensive User} gives a \textit{FORCE\_START} response to the critical utterance, then \colorbox{BurntOrange}{[ Music ]} provides a \textit{GENERIC} prompt to steer the conversation in a new direction. \\
    
    \specialrule{.01em}{.1em}{.1em} 
    
    11 & \textbf{User:} what do you find interesting & None &  The user gives an unexpected response. \\ [0.4ex] 
    
    & \textbf{Bot:} I like the violin, but I'm more of a classical music player. \colorbox{LimeGreen}{[ Neural Fallback ]} I remember you mentioned {Neo}. Would you like to talk more about it? \colorbox{SkyBlue}{[ Wiki ]} & Neo (The Matrix) &  \colorbox{LimeGreen}{Neural Fallback} gives a GPT-2-generated response, which is selected as no other RG (except \colorbox{Thistle}{Fallback}) has a response. \colorbox{SkyBlue}{Wiki} gives a \textit{CONTEXTUAL} prompt. \\
    
    \specialrule{.01em}{.1em}{.1em} 
    
    12 & \textbf{User:} sure & Neo (The Matrix) & \\[0.4ex] 
    
    & \textbf{Bot:} Neo has got into so many adventures! What are some adventures that you've enjoyed? \colorbox{SkyBlue}{[ Wiki ]} & Neo (The Matrix) & \colorbox{SkyBlue}{Wiki} asks an open-ended question (specific to movie characters) about \textit{Neo (The Matrix)}.\\
    
    \specialrule{.01em}{.1em}{.1em} 
    
    13 & \textbf{User:} morpheus teaching jujitsu to neo & Neo (The Matrix) & \\[0.4ex] 
    
    & \textbf{Bot:} I liked that Neo and Trinity were able to rescue Morpheus from a building protected by armed guards and agents. Morpheus has done so many amazing things! What are some of your favorite Morpheus moments? \colorbox{SkyBlue}{[ Wiki ]} & Morpheus (The Matrix) & \colorbox{SkyBlue}{Wiki} gives a response (generated by the GPT-2 conversational paraphrasing model), sets the current entity to \textit{Morpheus (The Matrix)}, and asks another open-ended question. \\
    
    \specialrule{.01em}{.1em}{.1em} 
    
    14 & \textbf{User:} i want to stop talking  & None & We detect \textit{stop intent}; the conversation ends. \\ 
  
    \specialrule{.10em}{.1em}{.1em}

 \caption{An example dialogue. To respect users' privacy, this is not a real user conversation.}

 \label{table:example_dialogue}
\end{longtable}

} 

\normalsize

\section{Dialogue Management}
\label{sec:dialogue_management}

Our Dialogue Manager handles the high-level logic of tracking which topics we are discussing with the user, and which responses (and prompts) should be used to form the bot's utterances.

\subsection{Navigational Intent Classifier}
\label{sec:navigational_intent}

A user has \textit{navigational intent} when they are indicating that they do (\textit{positive}) or do not (\textit{negative}) want to talk about a particular topic.
Users might give navigational intent while specifying the topic (\textit{can we talk about minecraft}, \textit{stop talking about minecraft}), or referring to the current topic (\textit{let's discuss this more}, \textit{could you change the subject}), or referring to no topic (\textit{alexa can we talk}, \textit{i don't want to chat any more}). 
Users sometimes give positive and negative navigational intent in the same utterance (\textit{i don't want to talk about movies any more let's chat about you}).
To recognize navigational intent, we use manually-constructed regexes, as they are quite high precision.

\subsection{Entity Tracker}
\label{sec:entity_tracker}
For our response generators to work together to discuss different topics smoothly, we must track which entities we are currently discussing, which we have finished discussing, and possible entities to discuss in the future.
This is the role of the \textit{entity tracker}.
We assume that at any point in the conversation, there is one \textit{current entity}, which is either a Wikipedia entity (see \secref{sec:entity_linker}) or None (if we're discussing something that does not have a Wikipedia article, such as \tabref{table:example_dialogue} Turn 3).\footnote{In the future, we plan to generalize the entity tracker to track non-Wikipedia discussion topics.}
The current entity is updated at most three times per turn (see Figure \ref{fig:overview_diagram}): 
\begin{enumerate}
\item \textbf{After analyzing the user's utterance}.
The entity tracker uses the entity linker's output, which is a priority-ordered list of possible entities mentioned by the user on this turn, along with their scores (see \secref{sec:entity_linker} for details.)  
If the user expressed negative navigational intent towards the current entity, it is rejected. 
If the user expressed positive navigational intent towards some topic, we search inside the topic slot in their utterance; the highest-priority entity with score over a low threshold (1,000) is chosen as current entity.
If there is a particular type of entity we expect the user to mention on this turn (e.g. if the bot asked \textit{What's your favorite movie?}) and there is an entity with the expected Wikidata category (e.g. \textit{film}) with score over a low threshold (1,000), it is chosen as current entity.
Otherwise, if the entity linker has made a prediction with sufficiently high score (over 10,000), it becomes the current entity.
If none of these conditions are met, the current entity stays the same.
\item \textbf{After choosing the response}. When the RGs provide responses, each RG also specifies what the new current entity should be, if its response is selected by the priority ranker. 
We update the current entity to be whatever was provided by the selected RG.
\item \textbf{After choosing the prompt}. If we get a prompt, we update the current entity similarly.
\end{enumerate}

This system allows the user to initiate topics (e.g. the bot starts talking about cats if the user utterance is \textit{i want to talk about cats}), allows RGs to initiate topics (see Table \ref{table:example_dialogue}, Turn 4), allows multiple RGs to talk seamlessly about the same topic (see Table \ref{table:example_dialogue}, Turn 10), and allows RGs to signal when a topic should be finished (see Table \ref{table:example_dialogue}, Turn 7).

\subsection{Response Priority Ranking System} \label{sec:response_priority}

We use a priority system to decide which response generator's response should be selected on each turn. 
When generating responses, each RG provides one of the \textbf{response priorities} in \tabref{table:response_priority_ranking_system}.\footnote{In case of a tie, we tie-break using a manually-specified priority ordering of the RGs.}
This hierarchy supports the ability to preserve conversational continuity (\textit{STRONG\_CONTINUE}), while remaining responsive to the user's initiative (\textit{FORCE\_START}).
Though it is a relatively simple rule-based system, we have found it well-suited to our needs.
The priority levels are clear to understand, and make it easy to modify behavior.
By avoiding a centralized response-choosing module, our design allows RGs to decide themselves whether or not they should respond, and whether their response is high quality.
This makes it easier for multiple people to work on different RGs, each with self-contained logic.
Lastly, if one RG encounters an error, timeout, or inability to find relevant content, the other RGs provide alternatives.

\begin{table*}[t]
  \begin{center}
    \small
    \begin{tabular}{ll} 
        \textbf{Response Priority} & \textbf{Meaning} \\
        \hline
        \textit{FORCE\_START} & This inactive RG should take control (e.g., \tabref{table:example_dialogue}, Turn 4), \\
        & or override, such as handling offensive user utterances (e.g., \tabref{table:example_dialogue}, Turn 10).\\
        \hline
        \textit{STRONG\_CONTINUE} & This active RG can continue the conversation with a good \\
        & next response (e.g., \tabref{table:example_dialogue}, Turn 2). Only a \textit{FORCE\_START} can override it. \\
        \hline
        \textit{CAN\_START} & This inactive RG can potentially take control (e.g., \tabref{table:example_dialogue}, Turn 8), \\
        & but should not interrupt a \textit{STRONG\_CONTINUE}. \\
        \hline
        \textit{WEAK\_CONTINUE} & This active RG can continue the conversation but its next response is of\\ 
        & poorer quality. It should be overridden by any available \textit{CAN\_START}s (or higher). \\
        \hline
        \textit{UNIVERSAL\_FALLBACK} & Only used by Fallback and Neural Fallback RGs (e.g., \secref{sec:rgs} and \tabref{table:example_dialogue}, Turn 11)\\ 
    \end{tabular}
    \normalsize
    \end{center}
    \caption{Response Priorities (ordered by descending importance)}
    \label{table:response_priority_ranking_system}
\end{table*}

\subsection{Response-and-Prompt System}
\label{sec:response_and_prompt}

As described in \secref{sec:system_overview}, on some turns the bot utterance consists of a \textbf{response} from one RG, followed by a \textbf{prompt} from another RG.
This system is useful when the responding RG can handle the user's current utterance, but is unable to take the conversation forward (see Table \ref{table:example_dialogue}, Turn 10) or when the responding RG has finished talking about one topic, and another RG is needed to supply a change of topic (see Table \ref{table:example_dialogue}, Turn 7). The response-and-prompt system makes it easy to always supply the user with a strong path forward in the conversation (e.g. by asking the user a question).

\subsection{Prompt Priority Sampling System}
\label{sec:prompt_priority}

While we use a deterministic ranking system to choose the highest-priority response (\secref{sec:response_priority}), \textit{prompts} often represent changes of topic, which are less restricted by context, and (in human-human conversations) tend to have a degree of randomness.
Thus, we use a priority \textit{sampling} system to select a prompt.
When generating prompts, each RG supplies one of the \textbf{prompt priorities} in \tabref{table:prompt_priority_ranking_system}.
  
\begin{table*}[t]
  \begin{center}
    \small
    \begin{tabular}{ ll } 
        \textbf{Prompt Priority} & \textbf{Meaning} \\
        \hline
        \textit{FORCE\_START} & This RG should take control. This is mainly used for scripted transitions (e.g., \tabref{table:example_dialogue}, Turn 2).  \\
        \hline
        \textit{CURRENT\_TOPIC} & This RG has a prompt that talks about the current entity (see Section 3.2 and \tabref{table:example_dialogue}, Turn 9). \\
        \hline
        \textit{CONTEXTUAL} & This RG has a prompt that does not talk about the current entity, but that is conditioned on \\ & the
        conversation history, e.g. referring to a previous topic (e.g., \tabref{table:example_dialogue}, Turn 11). \\ \hline
        \textit{GENERIC} & This RG has a prompt that is not conditioned on the conversation so far (e.g., \tabref{table:example_dialogue}, Turn 7). \\ 
    \end{tabular}
    \normalsize
    \end{center}
    \caption{Prompt Priorities}
    \label{table:prompt_priority_ranking_system}
\end{table*}
  
Under the Priority Sampling module, if a \textit{FORCE\_START} prompt is supplied, we choose it.
Otherwise, we sample from a manually-specified distribution over the remaining priorities, masking out any that are not present on this turn.
The distribution is biased towards maintaining continuity of discussion (\textit{CURRENT\_TOPIC} $\gg$ \textit{CONTEXTUAL} $>$ \textit{GENERIC}).
Then, among the RGs that produced a prompt of the sampled priority, we sample one prompt, using a manually specified distribution over the RGs.
This system allows us to specify scripted transitions when desired, and to provide variety via randomness, while still enabling us to tune the likelihood of changing topic, which is an important controllable parameter in chit-chat conversations \citep{see2019makes}.

\section{NLP Pipeline}
\label{sec:nlp_pipeline}
The NLP Pipeline is run at the start of every turn (see \figref{fig:overview_diagram}), and contains modules that annotate the user's utterance with information that is useful for other parts of the bot.

\subsection{CoreNLP}
\label{sec:corenlp}

On each turn of the conversation, we annotate the the user's utterance using the Stanford CoreNLP toolkit \citep{manning-EtAl:2014:P14-5}, which runs on a remote EC2 module with CPU only.
We use the following CoreNLP annotators: tokenization, sentence splitting, part-of-speech tagging, lemmatization, named entity recognition, constituency parsing, dependency parsing, coreference resolution, and sentiment analysis.
Due to the format of the user utterances (lowercase with no punctuation), we use the caseless models\footnote{\url{https://stanfordnlp.github.io/CoreNLP/caseless.html}} for part-of-speech tagging, constituency parsing and named entity recognition.

\subsection{Dialogue Act Classifier}
\label{sec:dialogue_act_classifier}

Dialogue acts can support understanding of user intent \citep{stolcke2000dialogue}, and have been successfully employed in previous Alexa Prize socialbots \citep{yu2019gunrock}.
To build a dialogue act classifier, we finetuned the HuggingFace implementation \citep{huggingface} of a BERT-based classification model \citep{bert} on the MIDAS dataset \citep{yu2019midas}. 
The dataset contains 12,894 examples, where each example is a bot utterance,\footnote{The bot utterance is included because it contains context essential to understand the user utterance \citep{yu2019midas}. 
For instance, the user utterance `tiger king' is an \textit{opinion}  when in response to `What is the best show?' and a \textit{statement} when in response to `What is the last show you watched?'.} the user's response to that utterance, and the user's dialogue act.\footnote{To better fit our needs, we modified the label space as described in \secref{sec:dialogue_act_label_modifications}.}
The dataset was collected by Gunrock \citep{yu2019gunrock}, the winner of the 2018 Alexa Prize competition. 
Unlike other dialogue act datasets, such as SWBD-DAMSL \citep{jurafsky1997switch}, which are designed for human-human dialogue, the MIDAS annotation schema was specifically designed for human-chatbot dialogue. 

\begin{table}
\begin{center}
 \begin{tabular}{lcccc} 
 \hline
 
 \textbf{Training Regime} & \textbf{\# MIDAS}  & \multicolumn{2}{c}{\textbf{Chirpy Training Set}}  & \textbf{Chirpy Test} \\ 
   &  \textbf{Training Set} & \textbf{\# Silver} & \textbf{\# Gold} &  \textbf{Set Micro-F1} \\[0.5ex]
 \hline
MIDAS (baseline) & 10,090 & 0 & 0 & 0.53 \\
MIDAS+self-training ($\tau = 0.95$) & 10,090 & 41,152 & 0 & 0.54 \\
MIDAS+self-training ($\tau = 0.75$) & 10,090 & 62,150 & 0 & 0.54 \\
MIDAS+supervised & 10,090 & 0 & 2,407 & \textbf{0.81} \\
 \hline
\end{tabular}
\end{center}
\caption{Performance of our Dialogue Act model under different training regimes.}
\label{da-performance}
\end{table}

Though this baseline model achieved a micro-average F1-score of 0.78 on the MIDAS test set, we wished to evaluate its performance in our \textit{own} bot's conversational setting.
We hand-labeled a `Chirpy' test set containing 602 examples from our bot's conversations.
The same baseline model achieved only 0.53 on this test set (see Table \ref{da-performance}). 
We suspect the performance drop is due to the distributional difference between the utterances generated by our bot and by Gunrock. 
To improve performance on our data, we experimented with self-training
\citep{mcclosky-etal-2006-effective}.
Using the baseline model, we labeled a large number of unlabeled examples from our own bot's conversations.
Examples whose label was predicted with a confidence score greater than a threshold $\tau$ were added to our training set.
Using $\tau = 0.75$ and $\tau = 0.95$ added 62,150 and 42,152 silver-labeled training examples, respectively.
After training on these expanded datasets, we re-evaluated on our own test set.
The inclusion of the silver-labeled data did not substantially boost performance (see \tabref{da-performance}).
Finally, we turned to supervised training, and hand-labeled an additional 2,407 examples from our own bot's conversations (procedure described in \secref{sec:dialogue_act_labelling}).
After training on the MIDAS data and this data, we achieved a much higher micro-F1 of 0.81 on the Chirpy test set.

In our bot, we run the Dialogue Act classifier on an EC2 machine with one NVIDIA T4 Tensor Core GPU, annotating every user utterance in the conversation.
We find that its accuracy is best on classes with low variance in user utterances, such as \textit{positive answer}, while classes with high variance, such as \textit{statement}, are more difficult.
However, even for the low variance classes, the classifier's labels are very useful -- we are able to achieve much higher recall in recognizing \textit{positive answer} and \textit{negative answer} by using the classifier's labels, compared to regexes or word lists.

\subsection{Question Classifier}
Users often spontaneously ask factual questions, personal questions, follow-up questions, and even questions unrelated to the current topic. 
Recognizing and answering these questions is important, particularly for user initiative, but is also non-trivial, as user utterances do not contain punctuation.

To recognize questions, we initially used the Dialogue Act classifier's labels (which include question types like \textit{factual question} and \textit{open-ended question}). 
However, this did not work well; the classifier seemed to condition too much on the bot utterance preceding the user utterance -- which is less useful for recognizing questions than other dialogue acts.
Instead, we fine-tuned a RoBERTa model \citep{roberta, huggingface} on an simplified version of the Dialogue Act training data, framing the task as binary classification, conditioned only on the user utterance.
This model achieved an F1-score of 0.92 and improved the reliability of question detection. 

The classifier's labels are used to determine when certain RGs should respond -- for example, when the Evi RG (\secref{sec:simple_rgs}) should answer a factual question.
The labels are also useful for the neural generative models (Sections \ref{sec:neural_chat}, \ref{sec:wiki}, \ref{sec:neural_fallback}). 
We observe that the GPT-2-based models are much more likely to answer (rather than ignore) a user's question if a question mark is present. 
Thus, we use the classifier labels to determine when to append a question mark to the user utterance.

\subsection{Entity Linker}
\label{sec:entity_linker}

A key part of our high-coverage strategy (\secref{sec:intro}) is \textit{entity linking} -- detecting when the user is referring to an entity, and identifying the correct entity.
To obtain our pool of potential entities, we processed a dump\footnote{\url{https://dumps.wikimedia.org}} of English language Wikipedia.
For each article (i.e. each entity $E$), we collected (a) the \textit{pageview} (number of views in one month), and (b) the \textit{anchortext distribution} $P_{\text{anchortext}}(a|E)$. 

To compute the anchortext distribution for an entity $E$, we count the number of \textit{anchortexts} (i.e., strings, lowercased) that are used as hyperlinks to $E$ across Wikipedia (e.g., the entity Barack Obama may be referred to using the anchortexts \textit{barack obama}, \textit{obama}, or \textit{president obama}). Then:
\begin{equation}
P_{\text{anchortext}}(a|E) = \frac{\text{count}(\text{links from $a$ to $E$})}{\sum_{a' \in A(E)} \text{count}(\text{links from $a'$ to $E$})}
\end{equation}
where $A(E)$ is the set of all anchortexts that link to $E$.
We store each entity, along with its Wikipedia article, pageview, anchortext distribution, and Wikidata categories\footnote{For each entity, we collected all its ancestors via the \textit{instance of} and \textit{subclass of} relations. For people entities, we also used the \textit{occupation} relation.} in an ElasticSearch index. 

After we receive the user's utterance $u$, we assemble the set of candidate spans $S$. 
$S$ contains all $n$-grams in $u$ with $n \le 5$, excluding $n$-grams that consist only of stopwords.
We then query ElasticSearch to fetch all entities $E$ which have at least one span $s \in S$ among its anchortexts. 
To determine which entities the user is referring to, we wish to estimate $P(E|s)$, the likelihood that a span $s$ is referring to an entity $E$.
We model $P(E|s)$ as a Bayesian system:
\begin{equation}
P(E|s) \propto P(E) \times P(s|E).
\end{equation}
We assume that $P(E)$ is proportional to the pageview for the entity $E$, and $P(s|E)=P_{\text{anchortext}}(s|E)$.
Therefore, we define the $\text{score}(s,E)$ of a span $s$ and and entity $E$ to be:
\begin{equation}
\label{eqn:entity_linker_score}
\text{score}(s,E) = \text{pageview}(E) \times P_{\text{anchortext}}(s|E).
\end{equation}

The output of the entity linker is a priority-ordered list of $(s,E)$ pairs. 
The ordering is calculated using manually-curated rules and thresholds on the following features: 
(a) the score of $(s,E)$,
(b) the maximum unigram frequency\footnote{The maximum unigram frequency of $s$ is the frequency of the most common unigram inside $s$, computed using this unigram frequency list for spoken English: \url{http://ucrel.lancs.ac.uk/bncfreq/flists.html}} of $s$, 
(d) whether $E$ is in a Wikidata category that is expected for this turn\footnote{For example, if the bot asked \textit{What's your favorite movie?}, an expected Wikidata category is \textit{film}.},
(c) whether $s$ is contained inside any other linked span (priority is usually given to the larger span).
The output of the entity linker is primarily used by the entity tracker (\secref{sec:entity_tracker}) to identify the current entity under discussion.

\customparagraph{Limitations}
We found the entity linker to be one of the hardest components of our bot to build.
One difficulty is that our notion of an entity -- anything with a Wikipedia article (e.g. \textit{Cat} or \textit{Musical instrument} in \tabref{table:example_dialogue}) -- is much broader than the traditional definition of Named Entities (which is typically restricted to particular types, such as people and locations).
Our motivation in this definition was to enable high-coverage world knowledge by enabling any Wikipedia article to become a focus of discussion.
However, this made the entity linker's job much more difficult.
The need to detect an extremely broad range of entities, with no restriction to certain types, made it much more difficult to find a good precision/recall tradeoff, leading to both false positive and false negative problems in the bot.
In the future, we will need to develop better approaches for identifying our expanded notion of entities, or find a way to support high coverage of topics without relying as much on the entity linker.

\customparagraph{ASR Error Robustness}
As we do not have access to original user audio,
ASR errors are a major source of difficulty, particularly when they occur within entity names.
For example, if the user wants to talk about the film \textit{Ford v Ferrari}, but the ASR transcription is \textit{four v ferrari}, our entity linker will fail to identify the correct entity, as the span \textit{four v ferrari} is not among the anchortexts for the entity \textit{Ford v Ferarri}.
To address this, we adapted our entity linker to be robust to phonetically-similar spans and anchortexts; our method is similar to \cite{chen2018gunrock}.

First, we converted all Wikipedia entity anchortexts to their phoneme and metaphone representations (e.g., \textit{Harry Potter} to \texttt{`HH EH R IY P AA T ER'} and \texttt{`HRPTR'}) with a grapheme-to-phoneme tool\footnote{\url{https://pypi.org/project/g2p-en/}} and the double metaphone algorithm,\footnote{\url{https://pypi.org/project/metaphone/}} and indexed the mapping from anchortext phonemes to Wikipedia entities in ElasticSearch. 
When running the entity linker, we convert all spans $s \in S$ to their phonetic representations and query the ElasticSearch index, which returns a set of anchortexts $A_{\mathrm{phon}}$ that have similar phonetic representations to any of the spans queried.
This allows us to expand the candidate pool for each span $s$, from entities for which $s$ \textit{is} an anchortext, to entities for which $s$ is \textit{phonetically similar} to an anchortext. 
Finally, we redefine $P(s|E)$ as follows: for each anchortext $a\in A_{\mathrm{phon}}$, we start by finding its best-matching span $s^*(a) = \arg\max_{s\in S}\mathrm{sim}(s, a)$ where $\mathrm{sim}(\cdot, \cdot)$ is a phoneme similarity function\footnote{implemented on lists of phonemes with Python's \texttt{difflib.SequenceMatcher}} between 0 and 1; then, we filter out anchortexts that are phonetically too dissimilar to each span with a threshold of 0.8, resulting in a set of anchortexts for each span $A(s) = \{a | a \in A_{\mathrm{phon}}, s=s^*(a), \mathrm{sim}(a, s) \ge 0.8 \}$. Finally:
\begin{align}
    P(s|E) \propto &\left\{\begin{array}{ll}
    \max_{a\in A(s)}\textrm{count(links from $a$ to $E$)}\times \mathrm{sim}(s, a) & A(s) \ne \emptyset\\
    0 & \textrm{otherwise}
    \end{array}\right.
\end{align}
This definition of $P(s|E)$ replaces $P_{\text{anchortext}}(s|E)$ in Equation \eqref{eqn:entity_linker_score}.

\section{Response Generators}
\label{sec:rgs}
In this section, we describe our Response Generators (RGs). 
Additional minor RGs are described in \appref{sec:appendix_rgs}.
We also describe \textit{treelets} (\secref{sec:treelets}), a system we used to organize many of our RGs.

\subsection{Treelets: A System to Organize Dialogue Graphs}
\label{sec:treelets}

Many of our response generators rely on \textit{treelets}, a modular programming abstraction which represents a single node in a dialogue graph. 
The treelet system is largely based on dialogue trees \citep{eliza} and dialogue-frame-based systems such as GUS \citep{gus}. 
We define a treelet to be a small, 1-turn dialogue `tree' that manages all decisions necessary to produce a bot response given a user's utterance. This involves interpreting the user utterance, creating the bot's response, and specifying the treelet that should take control on the next turn. 

\begin{figure*}[t]
  \begin{center}
    \includegraphics[width=0.8\textwidth]{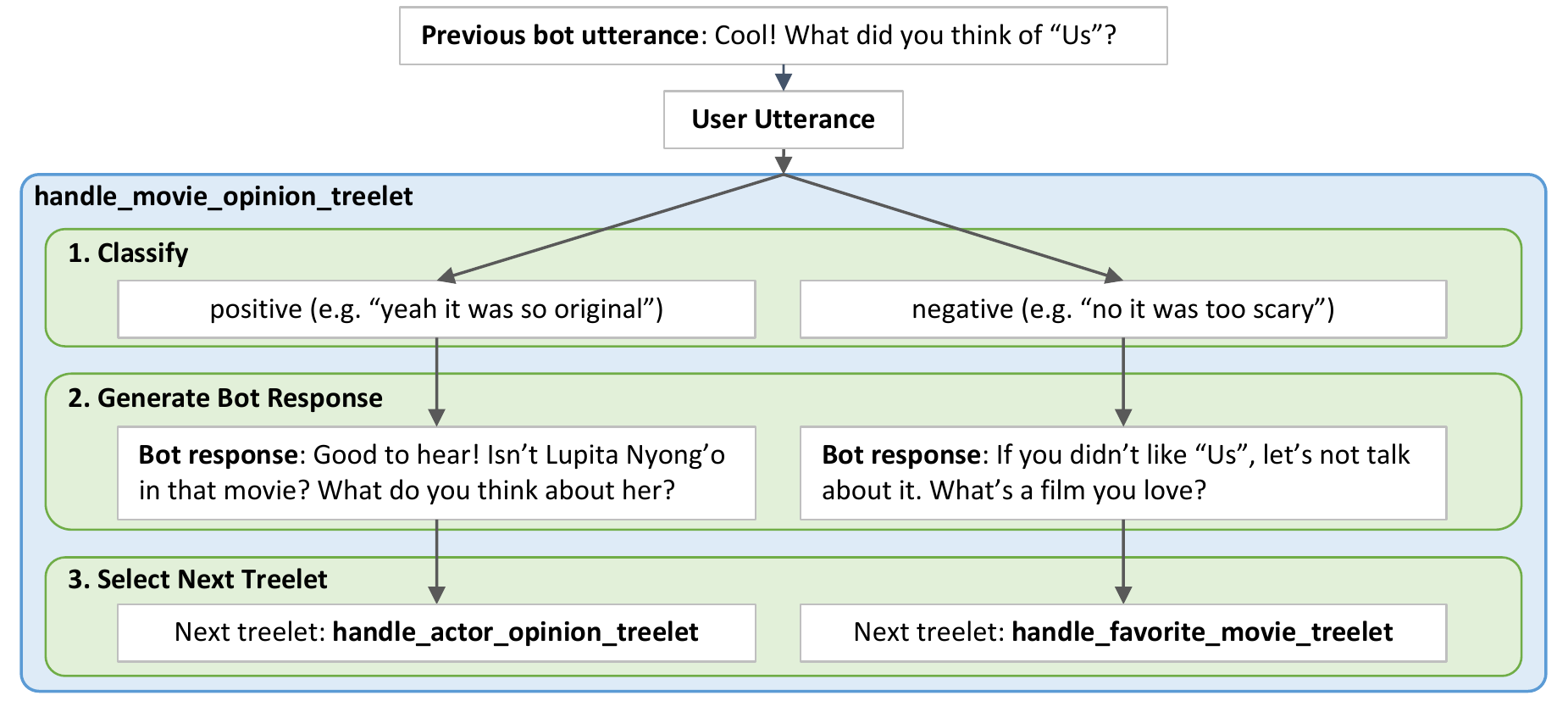}
    \caption{An example \textit{treelet} for the Movies RG.}
    \label{fig:example-treelet}
  \end{center}
\end{figure*}

Typically, a treelet performs three actions: (1) it classifies the user's utterance into one of several branches, (2) it produces an appropriate bot response for that branch, (3) it specifies the next treelet.
Treelets throughout our bot may classify user utterances by using regexes, outputs from our NLP pipeline (the dialogue act classifier is frequently used for this purpose), or changes in entity (e.g., if a treelet in the Movies RG detects that the current entity has changed to "food" after the user says "let's talk about food", the current Movies treelet may select a branch that returns no response). Bot responses may be handwritten or dynamically generated (we use both throughout our system). 
An example from the Movies RG is shown in Figure  \ref{fig:example-treelet}.

Like dialogue trees in general, treelets provide a well-controlled, predictable and easily interpretable conversation flow. 
From an engineering and implementation perspective, treelets have several advantages, such as allowing modular organization of code and dialogue, easily enabling cycles when desired (by having treelets point to each other with repeats or loops), and minimizing code duplication by allowing many treelets to point to the same successor.

\subsection{Neural Chat}
\label{sec:neural_chat}
The Neural Chat RG's goal is to empathetically discuss personal experiences and emotions with the user, using responses generated by a GPT-2-medium \citep{radford2019language} model finetuned on the EmpatheticDialogues dataset \citep{rashkin2019towards}.
The dataset consists of conversations between a \textit{speaker}, who describes an emotional personal experience, and a \textit{listener}, who responds empathetically to the speaker's story. 
Our model is trained in the listener role.

The Neural Chat RG has 7 discussion areas: current and recent activities, future activities, general activities, emotions, family members, living situation, and food.
A discussion begins by asking the user a \textbf{starter question} (e.g, \textit{What do you like to do to relax?} for the `general activities' area). 
Some starter questions are conditioned on the time of day (e.g. \textit{What did you have for breakfast/lunch/dinner today?} for the `food' area).
Starter questions can be asked as part of the launch sequence (\tabref{table:example_dialogue}, Turns 2 and 3), as generic changes of topic, (\textit{Do you have any plans for the weekend?}), or can be triggered contextually (\textit{You mentioned your boyfriend. How did you guys meet?}).
On each subsequent turn of the discussion, we generate 20 possible responses from the GPT-2 model using top-$p$ sampling with $p=0.9$ and temperature $0.7$. 
To provide a strong path forwards in the conversation, we generally choose a GPT-2 response containing a question.
However, if under a third of the sampled responses contain questions, we interpret this as an indication that the model is not confident in asking a question on this turn.
In this case, we choose a non-question and end the Neural Chat discussion.
Under this strategy, each Neural Chat discussion contains 2.75 bot utterances on average.

The model was finetuned using the HuggingFace ConvAI code\footnote{\url{https://github.com/huggingface/transfer-learning-conv-ai}} \citep{DBLP:journals/corr/abs-1901-08149} and is hosted on a GPU-enabled EC2 machine with one NVIDIA T4 Tensor Core GPU.
To keep latency low we truncate the conversational history supplied to the model, so that the total number of GPT-2 tokens is below 800.
Given that neural models have been shown to make poor use of longer conversational history \citep{sankar2019neural}, this truncation does not seem to be a limiting problem currently.

\begin{figure*}[t]
  \begin{center}
    \small
    \begin{tabular}{ |l|l| } 
        \hline
        \textbf{Strategy} & \textbf{Preamble} \\ \hline
        \rowcolor{timberwolf} \textbf{NO\_SHARE} & I wanted to check in with you.\\ \hline
        \rowcolor{babypink} \textbf{POS\_OTHERS} & I've noticed that a lot of people are feeling pretty positive today!\\ 
        \rowcolor{babypink} \textbf{POS\_BOT} & I wanted to say that I'm feeling pretty positive today!\\ 
        \rowcolor{babypink} \textbf{POS\_BOT\_STORY} & \textbf{POS\_BOT} + I just went for a walk outside, and it felt great to get some fresh air.\\ \hline
        \rowcolor{babyblue} \textbf{NEG\_OTHERS} & I've noticed that a lot of people are feeling kind of down recently.\\ 
        \rowcolor{babyblue} \textbf{NEG\_BOT} & I wanted to say that I've been feeling kind of down recently.\\ 
        \rowcolor{babyblue} \textbf{NEG\_BOT\_STORY} & \textbf{NEG\_BOT} + I've been missing my friends a lot and finding it hard to focus.\\ \hline
        \rowcolor{vanilla} \textbf{NEGOPT\_OTHERS} & \textbf{NEG\_OTHERS} + But I think its important to remember that things will get better.\\ 
        \rowcolor{vanilla} \textbf{NEGOPT\_BOT} & \textbf{NEG\_BOT} + But I think its important to remember that things will get better.\\ 
        \rowcolor{vanilla} \textbf{NEGOPT\_BOT\_STORY} & \textbf{NEGOPT\_BOT} + Just earlier today I took a walk outside and the fresh air \\
        \rowcolor{vanilla} & helped me get some perspective.\\ \hline
    \end{tabular}
    \normalsize
    \end{center}
    \caption{Strategies for the emotion-focused Neural Chat starter question. \textbf{POS}/\textbf{NEG}/\textbf{NEGOPT} refer to positive/negative/negative+optimistic emotion. \textbf{OTHERS}/\textbf{BOT} refer to whether the emotion is attributed to other people, or to the bot. \textbf{STORY} indicates that the bot shares a personal anecdote.}
    \label{table:neuralchat_emotion_questions}
\end{figure*}

\begin{figure*}[t]
  \begin{center}
    \includegraphics[width=0.8\textwidth]{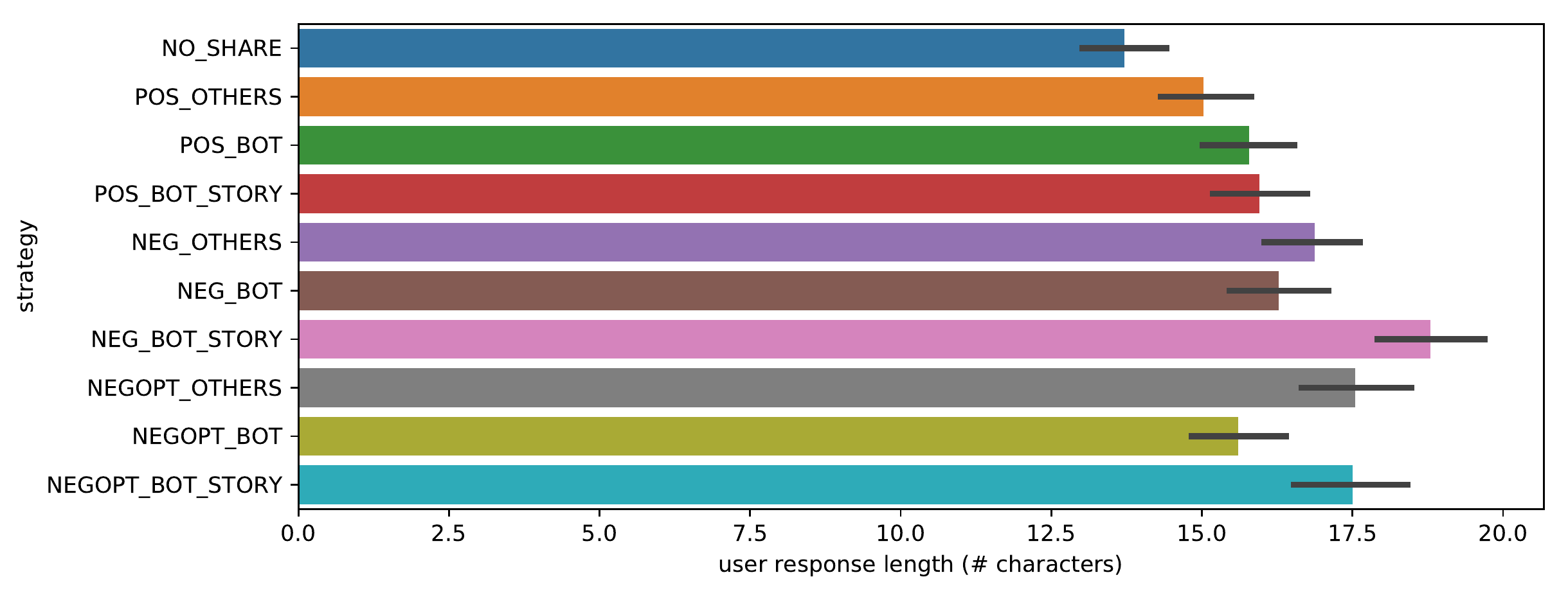}
    \caption{Effect of Neural Chat emotion-focused starter question strategies on user response length.}
    \label{fig:neuralchat_emotion_strategies}
  \end{center}
\end{figure*}

\customparagraph{Emotion-focused Conversations}
As part of our goal to provide an emotionally-engaging experience (\secref{sec:intro}), we would like to give users space to share their genuine feelings, then respond empathetically to them.
This is especially important during the Coronavirus pandemic (\secref{sec:coronavirus}), which is an emotionally challenging time for many.
Given our basic starter question \textit{I hope you don't mind me asking, how are you feeling?}, we tried several different preambles to precede the question (\tabref{table:neuralchat_emotion_questions}). 
\figref{fig:neuralchat_emotion_strategies} shows the effect of the different strategies on the length of the user's response.
We find that the basic \textbf{NO\_SHARE} strategy has the shortest average response length, indicating that the bot's emotional observations (whether about the bot or about other people) lead users to give more substantive responses.
Users tend to give longer responses when the bot expresses negative emotions (\textbf{NEG} and \textbf{NEGOPT}) than positive (\textbf{POS}) -- this may be because acknowledging negative emotions makes users feel more comfortable to answer the question honestly, rather than superficially (e.g. \textit{i'm fine}). 
Furthermore, adding a personal anecdote (\textbf{STORY}) to the negative bot emotions led to longer responses -- users may have responded more because the bot was more specific or relatable. 
For positive emotions (\textbf{POS}), users are more responsive when the bot attributes the positive emotion to itself (\textbf{BOT}), than to other people (\textbf{OTHERS}).
However, for negative emotions (\textbf{NEG} and \textbf{NEGOPT}), the opposite is true.
We also experimented with including the user's name in the starter question, but found that this made no difference to user response length.

\customparagraph{Discussion}
Our neural generative model has several recurring weaknesses which impact overall user experience. 
First, it frequently asks for already-provided information, asks nonsequitur questions, makes unfounded assumptions about the user, and confuses its own previous responses with the user's.
This demonstrates that incorporating commonsense reasoning is a priority in neural generation.
Second, while the model generally produces interesting and relevant responses to longer user utterances, it performs poorly when the user utterance is short or low-content (e.g. \textit{okay}, \textit{i don't know}, \textit{nothing}) -- probably because these utterances are unlike the much longer and contentful EmpatheticDialogues training data.
The model tends to respond to these with bland responses that further fail to drive the conversation to any interesting substance. 
This problem with short user responses is one reason why we focused on finding starter questions that lead to substantial user responses (\figref{fig:neuralchat_emotion_strategies}).

Due to these difficulties, most conversations with the GPT-2 model tend to fall apart after a few turns, as the bot will eventually ask a question that doesn't make sense, which will flummox the user. 
This is one reason why we designed the Neural Chat module around shorter sub-conversations.
However, overall, we are excited that neural generation is now able to interact successfully with real people, within certain constraints (such as keeping the discussion short, bookending it between handwritten starter questions and wrapup phrases, and providing a strong path forward through questions).

\subsection{Wiki}
\label{sec:wiki}
To support our goal of high-coverage world knowledge (\secref{sec:intro}), the Wiki RG uses Wikipedia articles as grounding to discuss any entity that interests the user.
Our goal is to allow the user to conversationally discover interesting information about the entity.

\customparagraph{Data} 
To prepare the Wikipedia data, we downloaded the most recent Wikipedia dump,\footnote{\url{https://dumps.wikimedia.org/backup-index.html}} processed it using MWParserFromHell\footnote{\url{https://mwparserfromhell.readthedocs.io/en/latest}} and Spark,\footnote{\url{https://spark.apache.org}} and uploaded it into an ElasticSearch index.
The Wiki RG can then query the ElasticSearch index to obtain the Wikipedia article for an entity.

\customparagraph{Behavior}
On each turn, if it's not already active, the Wiki RG can start to talk about the current entity (\secref{sec:entity_tracker}) by asking the user an \textbf{open ended question}, such as \textit{What do you find interesting about it?}. 
If the entity is in one of 25 commonly-encountered types (determined using Wikidata categories), such as books or foods, we use a more specific question, such as \textit{What did you think of BOOK\_ENTITY's story?} or \textit{I love trying out new flavor combinations. What do you like to have FOOD\_ENTITY with?}.
These questions are designed to elicit contentful user responses, which can be matched to specific sentences in the Wikipedia article using TF-IDF overlap.  
The RG also offers interesting facts (i.e. `TILs') scraped from the /r/todayilearned subreddit, if available.
If we have given enough TILs or we have no TIL left to offer, we will start suggesting sections of the Wikipedia article to the user. 
A short example Wiki interaction is shown in Turns 11-13 of Table \ref{table:example_dialogue}.

\customparagraph{Conversational Styling} 
We use this RG as a testbed for our conversational paraphrasing system. 
The system takes as input the truncated conversational history, and some knowledge context (either a TIL about the current entity, or an excerpt of the Wikipedia article, selected based on TF-IDF similarity to the user's response to an open-ended question). 
It outputs a conversational-sounding paraphrase of the knowledge context. 
The model was trained by finetuning a GPT-2-medium language model \citep{radford2019language} on a processed and filtered version of the TopicalChat dataset \citep{gopalakrishnan2019topical}. 
The paraphrases are generated using top-$p$ decoding with $p=0.75$ and temperature $\tau=0.9$, and we pick the one which has the highest unigram overlap with the knowledge context.

\customparagraph{Challenges} 
One major challenge while performing conversational styling is that the model sometimes produces \textbf{factually incorrect} or nonsensical conversational paraphrases. 
Another challenge is that integrating the paraphrasing model with the rest of the system requires \textbf{explicit directives} such as "continue talking about same knowledge piece", "pick another fact", "change entity" which the model currently does not produce. 
For instance, sometimes the generated paraphrase just asks a question or mentions an incomplete piece of information, with the expectation of completing it in the next turn.
Currently we apply some heuristics such as presence of \textit{Did you know ... ?} style questions or low unigram overlap to determine that the same snippet needs to be paraphrased again. 

More broadly, there are challenges around \textbf{interestingness of content}. The majority of content on Wikipedia isn't very interesting and social. While the TILs remedy that to some extent, finding interesting parts of raw text is still an open question and quite important in the open-domain conversational setting. 
Another major challenge is \textbf{content selection and discoverability}. The user doesn't know the extent of the knowledge that our system possesses for an entity. In a visual interface, the user can scroll through the article or look at a table of contents. While we partly remedy this by suggesting section titles to illustrate the kind of content we can talk about, a better system could perhaps understand what different parts of a Wikipedia article are talking about, and steer conversation in that direction.

\subsection{Opinion}
\label{sec:opinion}

Exchanging opinions is a core part of social chit-chat. 
To form a stronger sense of personality, and to seem more relatable, it is important that our bot can also express its opinions.
The Opinion RG's goal is to listen to users' opinions on certain topics, and reciprocate with its `own' opinions (sourced from Twitter) on those topics. 

\customparagraph{Data} 
To collect both positive and negative opinions, we queried a Twitter stream\footnote{https://developer.twitter.com/en/docs/tutorials/consuming-streaming-data} using a regex to collect tweets of the form \texttt{`i (love|like|admire|adore|hate|don't like|dislike) TOPIC because REASON'}, where \texttt{TOPIC} and \texttt{REASON} can be any text.
We collected ~900,000 tweets, which are stored on a Postgres table hosted on AWS Relational Database Service (RDS).
Of these, we manually whitelisted 1012 reasons across 109 popular topics. 
To avoid speaking inappropriately about sensitive topics, we only whitelist uncontroversial entities (such as animals, foods, books/movies/games, everyday experiences such as working from home, being sick, days of the week, etc.), and ensured that all reasons, including negative ones, are inoffensive and good-spirited.

\customparagraph{Behavior}
Currently, the Opinion RG activates when the user mentions one of the whitelisted entities (e.g. \tabref{table:example_dialogue}, Turn 8).
We ask whether the user likes the entity and classify their response using the CoreNLP sentiment classifier (\secref{sec:corenlp}). 
We then either agree or disagree with the user. 
If we disagree, we either ask the user for their reason for their opinion, or supply a reason why we disagree, and ask what they think of our reason.
Ultimately, we want the user to have a positive experience with our bot, so regardless of whether we disagree or agree with the user, we will ask the user their opinion on a related entity, and always agree with the user about the new entity. 
The conversation may end earlier, as we detect on each turn whether the user is still interested via their utterance length.
If the utterance contains less than 4 words, and it does not contain any of the `agreement' words (such as `same', `me too', etc.) we will hand off the conversation to another RG.
Even when the RG is not active, it keeps track of whether the user has already expressed an opinion on an entity, by applying a regex similar to that applied to the tweets.

\customparagraph{Agreement Policies}
Disagreement is an unavoidable part of human-human conversations, and we hypothesize that occasional disagreement is necessary in order for our bot to have a convincing and individual personality.
To test this, we implemented three policies (full details in Appendix \ref{sec:opinion_policies_details}):
(i)~\texttt{ALWAYS\_AGREE} -- we always agree with the user's sentiment on the entity;
(ii)~\texttt{LISTEN\_FIRST\_DISAGREE} -- first we ask the user's reason for liking/disliking the entity, then we offer our reason for disagreeing with their sentiment; and
(iii)~\texttt{CONVINCED\_AGREE} -- we initially disagree with the user's sentiment on the entity, but after the user gives their reason for liking/disliking the entity, we switch our sentiment to match the user's (i.e. we are convinced by the user).
To evaluate the policies, we ask the user \textit{Would you like to continue sharing opinions?} and interpret the desire to continue is an indication of a successful policy. 
\tabref{tab:cont_rate_opinion} shows that users prefer \texttt{ALWAYS\_AGREE} and \texttt{LISTEN\_FIRST\_DISAGREE} over \texttt{CONVINCED\_AGREE}, and all policies have high continuation rates, suggesting that disagreement can be a positive and stimulating part of a conversation, but that the manner and delivery of the disagreement is an important factor.

\begin{table}
    \centering
    \begin{tabular}{lrr}
    \hline
     \textbf{Policy Name}                    &   \textbf{Continuation Rate} &         \textbf{CI} \\
    \hline
     \texttt{CONVINCED\_AGREE} &     0.526829 & 0.0348712  \\
     \texttt{ALWAYS\_AGREE}              &     0.586638 & 0.0086009 \\
     \texttt{LISTEN\_FIRST\_DISAGREE}  &     0.587045 & 0.0127898  \\
    \hline
    \end{tabular}
    \caption{Continuation rate for each agreement policy. 
    The Confidence Intervals (CI) differ due to different sample sizes (ALWAYS\_AGREE receives 0.5 of traffic, LISTEN\_FIRST\_DISAGREE receives 0.3, CONVINCED\_AGREE receives 0.2).}
    \label{tab:cont_rate_opinion}
\end{table}

\subsection{Movies}
\label{sec:movies}

The Movies RG is designed to deliver a high-quality scripted conversation about a movie the user specifies, using information drawn from the Alexa Knowledge Graph.\footnote{The Alexa Knowledge Graph is an Amazon-internal resource; our team was given access to parts of it.} Currently, the RG is activated when the user asks to talk about movies, mentions a movie keyword (such as \textit{movies} or \textit{film}) or talks about any movie-related entity (e.g. \textit{Saving Private Ryan}, \textit{Meryl Streep}, \textit{the Coen brothers}, etc.). 
Once activated, the RG typically asks the user to name a movie, asks the user's opinion on it, gives a fun fact about the movie, asks the user their opinion on an actor in the movie, then asks the user if they've seen a different movie featuring that actor (See Turns 4-7 in Table \ref{table:example_dialogue}). The RG uses treelets (\secref{sec:treelets}) to organize the dialogue graph, hand-written templates to form the bot utterances, and a mixture of regexes and the CoreNLP sentiment classifier (\secref{sec:corenlp}) to classify the user's responses.

The primary weakness of this RG is that, as a scripted dialogue graph, it does not offer very high user initiative (one of our design goals -- \secref{sec:intro}).
However, this RG was important especially early in the competition when our more flexible RGs were still under development, and we needed more content. 
Another difficulty we faced was the latency of the Alexa Knowledge Graph, which was sufficiently slow that we were limited to one query per turn; this limited the scope of interesting information that we could pull about an entity and heavily influenced the design of our dialogue tree.

\subsection{Music}

Similar to the Movies RG, the Music RG is designed to deliver scripted conversations about musical entities that the user specify. The RG is activated when a musician/band or a music keyword (such as \textit{music} or \textit{songs}) is mentioned. Once activated, the Music RG engages in a conversation specific to the type of the musical entity that was mentioned. Unlike the Movies RG, the Music RG has a randomized internal prompting system that allows the conversation to be centered around music even when a scripted conversation is exhausted for a specific entity. For example, after the Music RG goes until the end of a scripted conversation for a musician, it can ask for an internal prompt, and start a conversation about musical instruments, songs, or music in general. The randomized nature of the internal prompting system makes the conversation more flexible, and mitigates some of the weaknesses of scripted conversations mentioned in \secref{sec:movies}.

\subsection{Neural Fallback}
\label{sec:neural_fallback}
Our Fallback RG's responses -- e.g., \textit{Sorry, I'm not sure how to answer that} (\secref{sec:simple_rgs}) -- are a poor user experience, making the user feel ignored and not understood. 
The Neural Fallback RG aims to generate a better fallback response using our GPT-2 EmpatheticDialogues model (\secref{sec:neural_chat}) -- to be used only if every other RG (excluding Fallback) has no response.
If the neural fallback response is chosen, another RG immediately produces a prompt to move the conversation in another direction.
After some filtering (e.g. removing responses that ask questions or give advice), the neural fallbacks can work well as a way to better acknowledge and show understanding of what the user said, such as on Turn 11 of Table \ref{table:example_dialogue}. 
A remaining issue is latency -- generating from the GPT-2 model is typically the slowest component in the turn, which is a poor tradeoff if we don't use the neural fallback.

\subsection{Categories}
\label{sec:categories}

The Categories RG was originally designed to ask handwritten questions about certain categories; for example, \textit{Where's a place you would love to visit?} for the `travel' category.
These questions may be asked when the current topic is `travel', or used as generic changes of topic (\tabref{table:example_dialogue}, Turn 7).
The goal is for the user to name an entity (e.g. \textit{japan}) that can form the basis for an interesting discussion (e.g. with the Wiki or Opinion RGs). 
However, we found that repeatedly asking users to think of entities led to decision fatigue, with many users failing to think of an entity.\footnote{If the user does not name a new entity, we respond either with a handwritten acknowledgment and new question (if the user said \textit{I don't know} or similar), or with the GPT-2 model (\secref{sec:neural_fallback}).} 
As alternatives to the \texttt{QUESTION} strategy, we experimented with two other strategies: \texttt{STATEMENT}, in which the bot just makes an observation about a relevant entity (e.g. \textit{Mexico is one of my favorite places. I love the food and beaches!}), and \texttt{STATEMENT+QUESTION}, which combines the other two strategies.
\tabref{tab:initiative_experiment} shows that the statement followed by a question elicited the most new entities.
This may be because the statement gives users an example, and takes the focus off the user for a moment, before prompting them with a question. 
This is a more natural, mixed-initiative experience than simply asking a question.

\begin{table}
    \centering
    \begin{tabular}{lrr}
    \hline
     \textbf{Strategy}                    &  \textbf{Proportion of Turns with New User Entities} & \textbf{CI}\\
    \hline
     \texttt{STATEMENT} & 0.272 & 0.012  \\
     \texttt{QUESTION}  &  0.264  & 0.027 \\
     \texttt{STATEMENT+QUESTION}  & 0.328 & 0.016  \\
    \hline
    \end{tabular}
    \caption{Rate at which users suggest new entities, for different strategies in the Categories RG. The entities are extracted using our Entity Linker (see Section \ref{sec:entity_linker}). (CI: Confidence Interval)}
    \label{tab:initiative_experiment}
\end{table}

\subsection{Offensive User}
\label{sec:offensive_user_rg}

Users sometimes give offensive or critical utterances, and it is important for our bot to handle these appropriately \citep{curry2018metoo, curry2019crowd}.
Unsurprisingly, there is an inverse relationship between the presence of offensive user utterances in a conversation and the conversation rating (\figref{fig:rg_coef}). Our goal is to redirect the user away from making offensive comments, towards topics the bot can discuss.

\begin{table}
    \centering
    \begin{tabular}{lrr}
    \hline
     \textbf{Strategy}                                      &     \textbf{Re-offense Rate} & \textbf{Confidence Interval}                  \\
    \hline
     \texttt{WHY}                                           &     0.520           & $\pm$0.049         \\
     \texttt{WHY+NAME}                                      &     \textit{0.638}  & $\pm$\textit{0.07}  \\
     \texttt{AVOIDANCE}                                     &     0.554           & $\pm$0.049          \\
     \texttt{AVOIDANCE+NAME}                                &     0.391           & $\pm$0.061          \\
     \texttt{AVOIDANCE+PROMPT}                              &     0.583           & $\pm$0.047          \\
     \texttt{AVOIDANCE+NAME+PROMPT}                         &     \textbf{0.346}  & $\pm$\textbf{0.066} \\
     \texttt{COUNTER+PROMPT}                                &     0.567           & $\pm$0.042          \\
     \texttt{EMPATHETIC+PROMPT}                             &     0.461           & $\pm$0.046          \\
    \hline
    \end{tabular}

    \caption{Re-offense rates for different response strategies to offensive utterances. Italic and bold denote the worst and best performing, respectively.}

    \label{tab:offensive-user-configurations-stats}
\end{table}

On each turn, the Offensive User RG checks the user's utterance for offensive language using a blacklist of offensive phrases.\footnote{\url{https://www.freewebheaders.com/full-list-of-bad-words-banned-by-google/}. 
Our offensive classifier is also used by our RGs to check that externally-sourced content (e.g. news articles, Wikipedia articles, fun facts) are inoffensive.}
If the user's utterance is more critical than offensive, we respond with an apologetic strategy (see Turn 10 of Table \ref{table:example_dialogue}).
For offensive user utterances, we implemented two immediate response strategies: asking the user why they made the offensive remark (\texttt{WHY}); or politely avoiding the topic (\texttt{AVOIDANCE}).
In addition, for \texttt{AVOIDANCE}, we experimented immediately changing the topic by using a prompt in the same turn (\texttt{AVOIDANCE+PROMPT}). 
For each of these configurations, we experimented with mentioning the user's name (\texttt{NAME}), or not.
We also implemented the strategy \texttt{COUNTER+PROMPT}, inspired by \cite{brahnam2005strategies}, which directly confronts the user before changing topic, and \texttt{EMPATHETIC+PROMPT}, inspired by \cite{chin2020empathy}, which empathizes with the user before changing topic.
The full details can be found in \appref{app:offensive}.

\tabref{tab:offensive-user-configurations-stats} shows the effect of each strategy on re-offense rate (i.e., the probability that the user says another offensive utterance in the same conversation). 
We find that mentioning the user's name reduces the likelihood of re-offense when we use the avoidance strategy, but increases re-offense rate when we ask the user why they made an offensive remark. 
We hypothesize that by using their name, we motivate the user to defend themselves, which prolongs the offensive conversation.
We find that our \texttt{AVOIDANCE+NAME+PROMPT} method outperforms the empathetic method (\texttt{EMPATHETIC+PROMPT}) and the confrontation method (\texttt{COUNTER+PROMPT}). 

\newpage

\section{Analysis}
\label{sec:analysis}
\subsection{Relationship between Rating and Engagement}
\begin{figure*}[h]
\centering
\vspace*{-3mm}
\includegraphics[width=\textwidth]{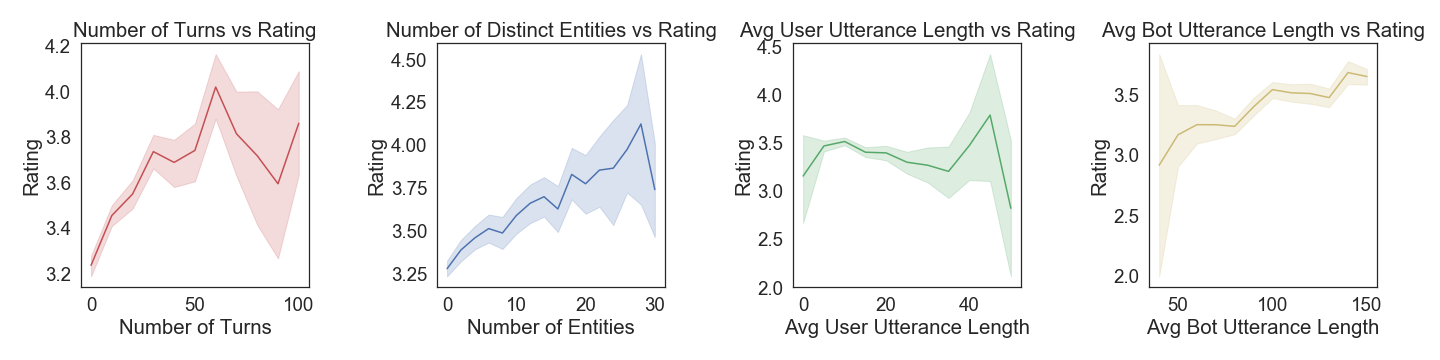}
\caption{Engagement metrics vs rating}
\vspace*{-3mm}
\label{fig:metrics_v_rating}
\end{figure*}
We measured four metrics of engagement: number of turns in the conversation, number of distinct entities discussed during the conversation, average length of the user's utterances, and average length of the bot's utterances. 
Figure \ref{fig:metrics_v_rating} shows that rating increases with number of turns and number of entities, but ultimately drops off. 
In an analysis of Alexa Prize bots, \cite{venkatesh2018evaluating} found that across all bots, conversation length was positively correlated with rating; however, one possible explanation for our result is that our bot has limited content  and at some point, the users become dissatisfied as their experience is no longer novel.

In an analysis of the NeurIPS ConvAI2 challenge, \cite{dinan2019second} found a positive relationship between user utterance length and rating. 
We expected a similar result, thinking more talkative users would be more actively engaged.
However,  Figure \ref{fig:metrics_v_rating} shows that rating increases with user utterance length until about 12 characters, and then decreases. 
Since many of our bot's questions encourage short answers (e.g. \textit{What's your favorite animal?}; \textit{Would you like to talk about science?}), and it is generally more difficult for our bot to correctly understand and handle longer answers,\footnote{As an exception, our neural generation models perform \textit{better} on longer user utterances; see \secref{sec:neural_chat}.} users who give longer answers may have a worse experience.
For this reason, the result shown may reflect the limitations of our bot, more than a user preference for giving shorter responses.

Average bot utterance length is positively correlated with average rating, 
with 
high variance in rating for shorter bot utterances. 
A confounding factor 
is that different response generators have varying average response lengths and relationship with user experience (\secref{sec:analysis_rgs}) -- e.g., the Offensive User RG tends to give short responses, and has a negative relationship with ratings.
Response generators giving longer responses tend to have positive or neutral relationships with rating. 
Therefore, this plot may more reflect the UX of our response generators than a user preference for longer responses. These results may also reflect the inherent noise in user Likert-scale ratings \citep{liang2020user}. 

\begin{figure}[t]
\centering
\parbox{6cm}{
\includegraphics[width=6cm]{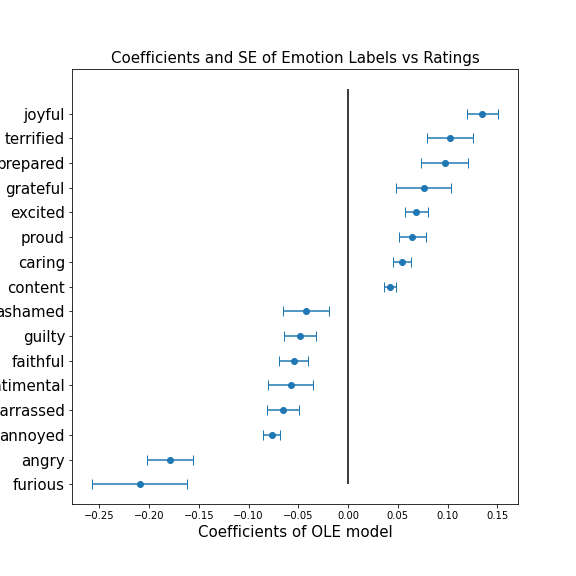}
\caption{Regression coefficients for Emotion vs Rating}
\label{fig:emotion_coef}}
\qquad
\begin{minipage}{6cm}
\includegraphics[width=6cm]{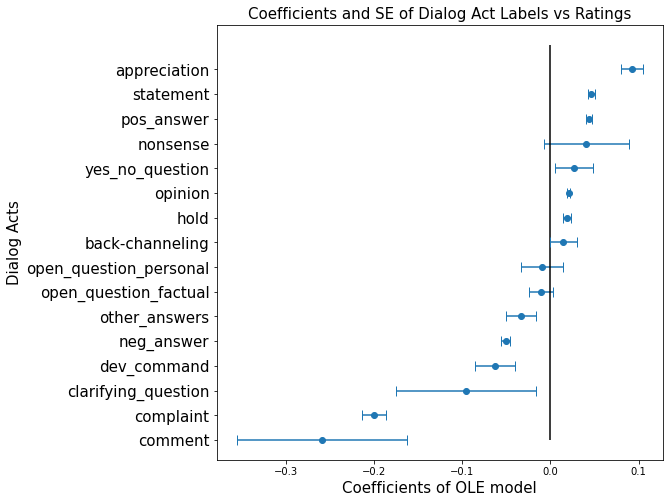}
\caption{Regression coefficients for Dialogue Act vs Rating}
\vspace*{-3mm}
\label{fig:da_coef}
\end{minipage}
\end{figure}
\begin{figure}[h]
\centering
\parbox{6cm}{
\includegraphics[width=6cm]{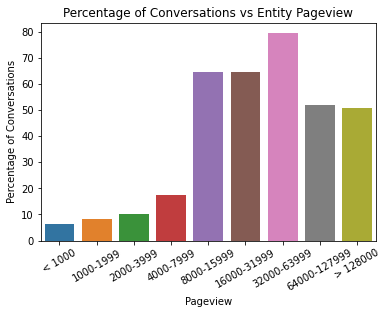}
\caption{Percentage of conversations in which users initiated discussion of entities with different popularity levels (pageview).}
\label{fig:entity_counts}
\label{fig:entity_pageview}}
\qquad
\begin{minipage}{6cm}
\includegraphics[width=6cm,height=4.5cm]{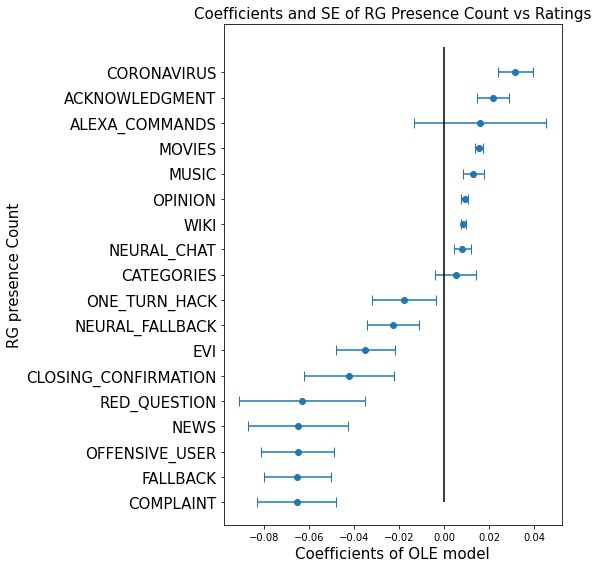}
\caption{Regression coefficients for Response Generator vs Rating. Launch RG is not included as it is in every conversation.}
\label{fig:rg_coef}
\end{minipage}
\end{figure}

\subsection{Relationship between Rating and User Dialogue Acts}

To understand how users' dialogue acts relate to our bot's performance, we applied a regression analysis, using the statsmodels \cite{seabold2010statsmodels} implementation of Ordinary Least Squares, to the distinct dialogue act classifier labels for all utterances of a conversation and the ultimate rating of that conversation. These results are shown in \tabref{fig:da_coef}. As we would expect, \textit{appreciation} is associated with higher ratings and \textit{complaint} with lower ratings. 

One of our design goals was having mixed-initiative dialogue. In general, dialogue acts associated with low user initiative, such as \textit{comment}, \textit{pos\_answer}, \textit{statement}, and \textit{back-channeling} were more positively associated with rating than dialogue acts associated with high user initiative, such as \textit{command}, \textit{open\_question\_opinion}, and \textit{open\_question\_factual}. A possible explanation for this is that users take more initiative when dissatisfied with the current conversational direction, for example by giving a command to change the topic. On the other hand, users giving yes-answers or back-channeling, are likely being compliant with the bot's direction, which may reflect greater overall satisfaction. It is possible that these results are more indicative of user satisfaction with our content than of a user preference for low vs high initiative.

\subsection{Entity Coverage}
\label{sec:analysis_entitytracker}

As part of our design goal to offer high coverage of topics (\secref{sec:intro}), our bot is capable of discussing any Wikipedia entity (\secref{sec:entity_tracker}), and discussed $7.5$ distinct entities on average per conversation.
To support user initiative and engage users, we designed our bot to be able to discuss both popular and lesser-known entities.
We regard the Wikipedia pageview (\secref{sec:entity_linker}) as a measure for an entity's popularity.
To measure users' desire to discuss less-common entities, \figref{fig:entity_counts} shows the percentage of conversations where users initiated discussion of an entity with different pageview levels. 
These counts do not include entities initiated by the bot. 
As the plot shows, a significant number of users wanted to discuss uncommon entities: in $8\%$ of our conversations, users initiated discussion of entities with fewer than 2000 views and $33\%$ of conversations covered at least one entity with fewer than 8000 views. Users who discussed rare entities with the bot appeared to have favorable experiences. Conversations with rare entities (fewer than 16000 pageviews) had an average rating of 3.88, while those without rare entities had an average rating of 3.64.  

To understand which entities had the greatest impact on user experience, we used the top 100 most frequent entities as features for a regression analysis, using an Ordinary Least Squares model. 
Of the 100 most popular entities, 15 had a statistically significant ($p \leq 0.05$) positive impact on rating. 
These include \textbf{animals} (`Cat', `Dog'), \textbf{movies} (`Film', `Frozen 2', `Onward (film)'), \textbf{food} (`Korean fried chicken', `Pizza', and `Ice cream'), and \textbf{video games} (`Minecraft', `Fortnite').

\subsection{Effectiveness of Response Generators}
\label{sec:analysis_rgs}

We performed a regression analysis on the relationship between response generator use and rating, using the number of turns each RG contributed as features. 
\figref{fig:rg_coef} shows a statistically significant positive relationship between rating and the Coronavirus, Acknowledgment, Movies, Opinion, and Wiki RGs, and a statistically significant negative relationship for Red Question, Complaint, Fallback, Neural Fallback, and Offensive User.
The Complaint and Offensive User results may be explained by the fact that users experiencing poor conversations may complain or be offensive, and conversely, some adversarial users deliberately engage negatively and then give poor ratings.
A possible cause for the negative Fallback and Neural Fallback results is that these RGs are used when no other RG has a high-quality response, so their use is likely correlated with a worse user experience. 
As we expected, RGs designed for general conversation had more positive coefficients. Of these RGs, those with more scripted content, i.e. Coronavirus, Acknowledgment, Movies, and Categories had larger positive coefficients than those with less, such as Opinion and Wiki. However, the most significant loss in performance occurs when the bot cannot answer contextually or has an adversarial user.

\section{Discussion and Future Work}
\label{sec:discussion}

\customparagraph{Full Stack NLP}
Most NLP research focuses on self-contained tasks. 
However, an open-domain socialbot, served to a diverse range of customers in widely different contexts, is by no means a self-contained task. 
Our socialbot is a tapestry of many such components, requiring a deep understanding of each component and how they should work together -- a setting we call Full Stack NLP.
Often the inputs and outputs of these components are inter-dependent, leading to cascading errors. 
We made many design choices which delay hard decisions in pipelines, and maximize information exchange between modules. 
Moving forward, the next avenue for advancing the state-of-the-art would be research on models which perform these tasks jointly and methods which enable training over multiple interdependent tasks with only a small amount of joint supervision. 

\customparagraph{Domain Shift} 
As a recurring problem, we found that many existing NLP resources didn't work well out-the-box.
The main reason for this is that the training data for these resources (typically non-conversational, longform, traditionally-formatted written text) is misaligned with our setting (conversational, shortform, uncased, punctuationless, spoken text).
However, a deeper reason is the constantly changing nature of dialogue agents themselves.
Even for an extremely related resource (the MIDAS dialogue model, developed for the Alexa Prize, \secref{sec:dialogue_act_classifier}), domain shift was a problem.
Recent advances in online- and meta-learning could provide a useful long term solution to this issue.

\customparagraph{Conflict and Intimacy}
Bot-human conversations are fundamentally different to human-human conversations. 
Users can be adversarial, deliberately testing the bot's boundaries. 
As socialbot designers, we are eager to avoid a disaster like Microsoft Tay, so we apply strict but overly simplistic methods to block off sensitive topics (Sections \ref{sec:opinion}, \ref{sec:offensive_user_rg}). 
However, this rules out sincere conversation about difficult topics. 
We observed that users are actually quite resilient to conflict, and can find disagreement stimulating (\secref{sec:opinion}). 
We also found that emotional intimacy is reciprocal -- users are more inclined to share their feelings after the bot has shared its own (\secref{sec:neural_chat}). 
Going forward, we should continue to take seriously the dangers of speaking inappropriately, but keep in mind the cost -- to engagement and to intimacy -- of not engaging in difficult topics.

\customparagraph{Initiative}
As part of our goal to support user initiative, we focused on asking users questions to find out which topics interested them.
However, this puts pressure on the user to think of a response, especially given the time constraints of Alexa devices.
Thus we found that our attempts to let the user take more initiative unfortunately led to decision fatigue. 
Separately, our ability to support user initiative was limited by our ability to answer followup questions, and to correctly understand long or unexpected user utterances. 
On balance, we found that asking the user open-ended questions about interesting topics was a good strategy -- easier to handle than spontaneous user questions, and less pressuring than asking users to name topics.
We see an opportunity for future work to build systems which listen more to the user's knowledge, rather than only providing knowledge.

\section*{Acknowledgments}
Thank you to Anna Goldie for her advice and guidance to the team.
Abigail See's work was supported by an unrestricted gift from Google LLC. We thank Amazon.com, Inc.\ for a grant partially supporting the work of the rest of the team.

\bibliography{ap_2020}
\bibliographystyle{acl_natbib}

\appendix

\section*{\Large Appendix}
\section{Additional RGs}
\label{sec:appendix_rgs}

\subsection{Coronavirus}
\label{sec:coronavirus}
Our bot has been available to users since early December 2019 -- thus covering the entire period of the COVID-19 pandemic so far. As the pandemic progressed, users requested to talk about the pandemic more and more, and showed more interest in discussing home life rather than outside activites. We created the Coronavirus RG to meet users' expectations that our bot should be fully aware of the unfolding situation, on both an informational and emotional level.

The RG is activated when the user mentions or requests to talk about the Coronavirus. 
To be mindful of the stress that the pandemic has placed on many people, we never initiate the topic ourselves. We then give a comforting message acknowledging the emotional effect of the pandemic, and give the user the option to hear up to 4 news updates. After that, we ask an uplifting open-ended question, and handle the user's response by generating a response from our GPT-2 model (\secref{sec:neural_chat}).
We then perform a scripted transition to the Movies RG\@. For example after we've read out a news headline:
\begin{quote}
    \textbf{User}: \textit{No thank you. I'm exhausted from listening to the news all the time.} \\
    \textbf{Bot}: \textit{No problem. I know this situation has been very stressful for many people; I’ve read online that it can be helpful to remind ourselves of small, good things that have happened to us lately. What’s something nice that’s happened to you in April?} \\
    \textbf{User}: \textit{Well it was my birthday!} \\
    \textbf{Bot}: \textit{That's great. Did you have fun?} \\
    $\ldots$
\end{quote}

\subsection{News}
The News RG provides the user with news stories that interest them, and discusses them. 
We set up a cron job to query the Washington Post API\footnote{An API call to scrape Washington Post news articles provided by Amazon Alexa.} and scrape several news-based subreddits\footnote{/r/News, /r/Sports, /r/Politics, /r/Futurology, /r/Science, /r/Technology, /r/WorldNews} every day, and place the data in an ElasticSearch index. 
When the user asks to talk about news, the News RG asks the user for a topic (e.g., \textit{Joe Biden}), if it wasn't already supplied.
The RG then queries the ElasticSearch index for news stories with the desired topic in the headline, selects the most recent one, reads out the headline to the user, and asks if they'd like to hear more. 
If accepted, we read out the first three sentences of the article.

Our original goal was to allow the user to ask \textit{}{follow-on questions} about the article, and to answer them with a \textbf{Neural Question Answering} model. We hoped this would help realize our design goals of conversational phrasing and enabling user initiative (\secref{sec:intro}).
To begin this process, the News RG would invite the user to ask questions.
We then used the SpaCy coreference resolution module \citep{spacy2} to decontextualize the user's question with respect to the last two utterances from the News RG. 
For example, \textit{how many votes did he win?} might be transformed to \textit{how many votes did Joe Biden win?} The decontextualized question, along with the entire news article, was then sent to a BERT-Large model \citep{bert} trained on the Stanford Question Answering 2.0 dataset \citep{squad2} by HuggingFace.\footnote{\url{https://github.com/huggingface/transformers}}
The model would output either a span in the article, or `no-answer' -- meaning the question cannot be answered by the 
provided article.\footnote{Since the article was often much larger than the maximum context size for BERT, we ran the model on chunks. Within each chunk, we discarded spans which were ranked lower than `no-answer', then merged the answers and re-ranked by confidence of the predictions.}

Unfortunately, in our internal testing, we found that this system had several substantial problems. First, errors in the coreference module were common, and would cascade to the QA module. Second, we found that users asked a very different distribution of questions, compared to the SQuAD training questions.
For example, users were likely to ask more open-ended or causal questions (e.g., \textit{what happened next?}, \textit{why did they do that?}). 
These are difficult for off-the-shelf QA models, which tend to excel in answering factoid-style questions. Third, users were likely to ask questions whose answers are not present in the news article. Though our model was trained on SQuAD 2.0 (which contains unanswerable questions), it would often choose an irrelevant answer that type-checks with the question, as \citet{jia2017adversarial} have also reported.
Even when the QA model correctly classified unanswerable questions, we would have needed to build a substantial open-domain question answering system to handle these questions. Overall, these problems made our system a poor and unreliable user experience; requiring more time and effort to fix than we had available.

\subsection{Other RGs}
\label{sec:simple_rgs}

\customparagraph{Launch} Handles the first few turns of the conversation (introducing the bot and learning the user's name). An example can be seen in Table \ref{table:example_dialogue}.

\customparagraph{Acknowledgment} When the user changes topic to a new entity, this RG uses the entity's membership in certain Wikidata categories to select a one-turn scripted acknowledgment (e.g. \textit{Oh yeah, I read ENTITY last year - I couldn't put it down!} if the entity is a book). This RG aims to give a natural and conversational acknowledgment that a new topic has been raised, before handing over to another RG (e.g. Wiki/Opinion/News) to discuss the entity in more depth. 

\customparagraph{Alexa Commands} Users often try to issue non-socialbot commands (such as playing music or adjusting smart home devices) to our socialbot. This RG detects such commands, informs the user that they're talking to a socialbot, and reminds them how they can exit.

\customparagraph{Closing Confirmation} Our bot stops the conversation when the user issues a command like \textit{stop} or \textit{exit}. However, users indicate a possible desire to exit through many other more ambiguous phrases (e.g., \textit{do you just keep talking}, \textit{what's happening}). This RG detects such cases using the \textit{closing} dialogue act label (\secref{sec:dialogue_act_classifier}) and regex templates, asks the user if they'd like to exit, and stops the conversation if so.

\customparagraph{Complaint} Provides an appropriate response when a user complaint is detected. This RG uses the Dialogue Act classifier's \textit{complaint} label to detect generic complaints, and regular expressions to detect misheard complaints (the user saying that Alexa misheard them), clarification complaints (the user saying that Alexa is not being clear), repetition complaints (the user saying that Alexa is repeating itself), and privacy complaints (the user saying that they don't want to share information). We wrote different responses for each type of complaint, to reflect understanding of the user's concerns.

\customparagraph{Fallback} Always provides a response (\textit{Sorry, I'm not sure how to answer that}) or prompt (\textit{So, what are you interested in?}) to be used when no other RG provides one.

\customparagraph{One-Turn Scripted Responses} Provides handwritten responses to common user utterances (e.g. \textit{help}, \textit{chat with me}, \textit{hello}) that can be handled in a single turn.

\customparagraph{Red Question} Detects if the user asks our bot a `red question' -- i.e., a question we are not permitted to answer, such as medical, legal, or financial advice -- and informs the user that we cannot answer.
To recognize these questions, we trained a multinomial logistic regression model on bag-of-words features, using data from the /r/AskDoctor, /r/financial\_advice, and /r/LegalAdvice subreddits.

\section{Tooling and Processes}
\label{sec:quality_and_monitoring}

\subsection{Dashboard}
\begin{figure}
    \centering
    \includegraphics[width=\linewidth]{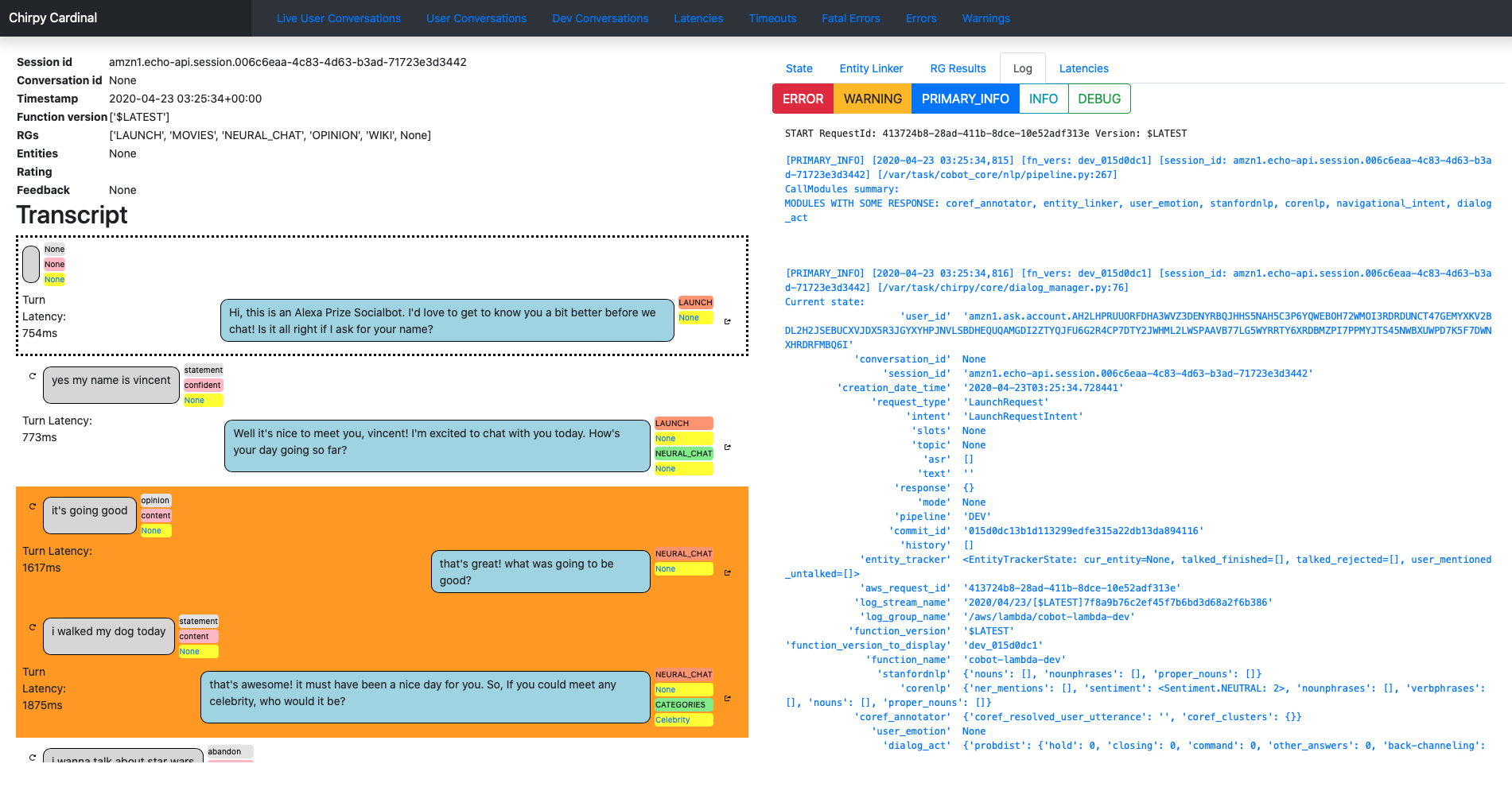}
    \caption{Screenshot of an example conversation (not with a real customer) in the dashboard. The tags next to each utterance are annotations from the bot. The background color of the utterance is the latency of that specific turn (white being normal and orange being slow). The pane on the right shows the logs for the turn.}
    \label{fig:dashboard_convo}
\end{figure}
We built a browser-based dashboard to provide ourselves with easy readable access to conversations and the associated metadata. 
The landing page shows aggregate rating statistics broken down by date and code version. 
The dashboard can filter conversations based on metadata such as number of turns, ratings, entities and RGs used. 
For each conversation, the dashboard displays important turn-level attributes, such as latency, entities, annotations, state information, RG results, and logs.
It can provide a link pointing to a specific turn, which is very useful for discussions and issue tracking. 
The dashboard can rerun the conversation with the current version of our bot, to quickly test if our local changes fixed the problem.
Aside from displaying conversations, the dashboard also has tabs to track errors and latencies, divided by severity level.
Easy accessibility and visibility of errors made us more aware and likely to fix these errors quickly. 

\subsection{Processes}
\customparagraph{Code Review} 
We realized early on that maintaining high code quality is important for maintainability and extensibility. 
We set up a circular code review process to ensure that any code we write is understandable by another team member and adheres to certain quality standards. 

\customparagraph{Integration Tests} 
We also instituted integration tests, to ensure that our bot maintains certain core functionality. 
We often found that some changes we made in one part of the bot had unexpected and damaging effects in another part of the bot; integration tests helped to catch these issues.

\customparagraph{Canary Testing} 
We had two versions of our bot -- \textbf{mainline}, which handled real customers, and \textbf{dev}, which we used for developing new features.
At first, new dev versions were solely tested by team members, before being pushed to mainline.
However, especially as the complexity of the bot grew, this method became insufficient to identify problems in new dev versions -- meaning that bugs were being discovered in mainline.
We set up a canary testing framework, which directs a controllable percentage (typically 10\%-50\%) of customer traffic to dev.
This was very useful in allowing us to tentatively test out new features with larger numbers of people, before deploying to all customers, thus protecting our ratings.

\customparagraph{UX Officer} 
Each week, we have a dedicated UX officer, whose primary responsibility is to monitor the conversations, identify problems, and get a sense of the strengths and weaknesses of the current system.
This person is also responsible for alerting other team members to things that need to be fixed, and communciating their overall findings to the rest of the team at the weekly meeting.
The role rotates every week so every team member has a chance to see the bot in action, and stay in touch with the overall user experience.

\customparagraph{Sprint Planning and Issue Tracking}
We use Jira to track issues to be fixed -- each is assigned to the person in charge of the relevant component.
We have a weekly sprint planning meeting where we prioritize the most important things to work on over the next week, and use Jira to track the sprint.

\section{Dialogue Act Classifier}
\subsection{Modifications to Label Space}
\label{sec:dialogue_act_label_modifications}
We modified this schema to better fit the needs of our bot, adopting 19 out of 23 dialogue act labels from MIDAS paper, and creating 5 new labels: \textit{correction}, \textit{clarification}, \textit{uncertain}, \textit{non-compliant}, and \textit{personal question} to support UX-enhancement features such as the ability to respond to clarifiying questions. We dropped the labels \textit{apology}, \textit{apology-response}, \textit{other}, and \textit{thanks} since there were very few (n $\leq$ 80) examples of them in the original dataset and we rarely observed these dialogue acts in our bot. 

\subsection{Labeling Procedure}
\label{sec:dialogue_act_labelling}
To create our gold-labeled dataset from our bot, we first determined which classes we most wanted to improve, based on per-class F1-Score for the baseline model and the new features we wanted to build. For example, since we wanted to improve our complaint handling, we prioritized this category. Next, we ran the baseline model on data from our bot to collect pseudo-labels. We randomly sampled 300 examples per label and then annotated whether the true label matched the predicted label. If not, we annotated what the correct label was. Using the pseudo-labels as a starting point increased efficiency, since the binary decision of "correct or incorrect" is much easier than the choice between 24 labels, and this method significantly reduced the number of non-binary decisions necessary. It also improved balance over classes, since it gave us greater control over the classes in the sample, and allowed us to prioritize certain categories. The result of training with gold-labeled examples is reported in Table \ref{da-performance}.

\section{Emotion classifier and analysis}
\label{sec:emotion_classifier}
\begin{figure}
\begin{center}
\includegraphics[width=0.9\textwidth]{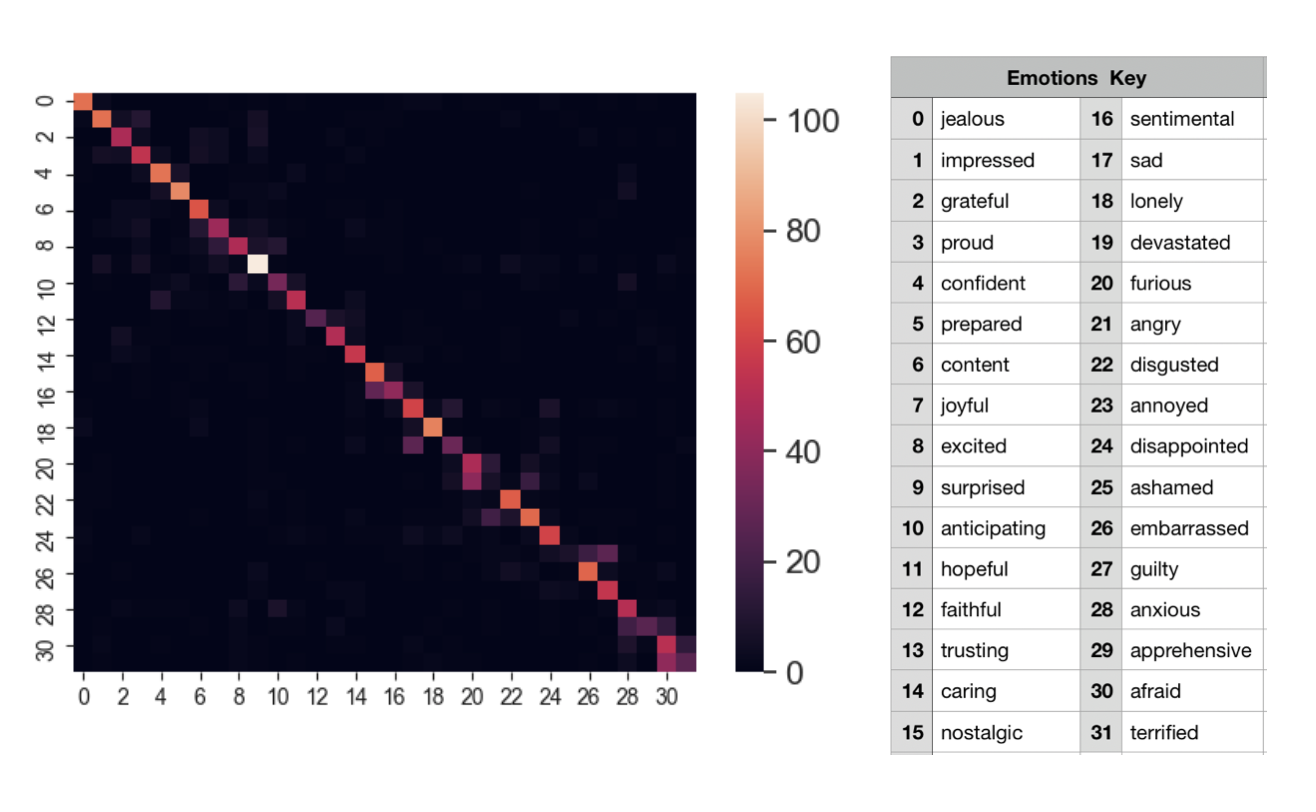}
\captionsetup{margin=0.5cm}
\caption{Confusion matrix for RoBERTa emotion classifier.}
\label{fig:confusion}
\end{center}
\end{figure}

In order to understand and analyze users' emotions, we finetuned a RoBERTa model \citep{roberta, huggingface} on the EmpatheticDialogues dataset \citep{rashkin2019towards}, which contains 24,850 examples broken into an 80-10-10 train-dev-test split. 
In particular, our training and test data consisted of the first utterance from each dialogue (as it is the only one with a label), along with its label (one of 32 fine-grained emotions, listed in  \figref{fig:confusion}).

The RoBERTa model achieves a top-1 accuracy of $61.5\%$ and an F1-score of $0.596$. 
However, many of the misclassifications are due to the model choosing a label very similar to the gold label.
For example, in the confusion matrix in \figref{fig:confusion}, we see that \textit{angry} is often misclassified as \textit{furious}, and \textit{terrified} as \textit{afraid}, among others. 
In contrast, the top-5 accuracy is $92\%$.

One difficulty in applying this classifier to our user utterances is domain shift.
The EmpatheticDialogues training utterances all describe a strongly emotional personal situation in complete written sentences, in a self-contained way (i.e., with no preceding context) -- for example, \textit{A recent job interview that I had made me feel very anxious because I felt like I didn't come prepared.} 
By contrast our user utterances are spoken, typically not complete sentences, require conversational context to understand, and encompass many different dialogue functions (such as giving commands, answering questions, choosing topics, greeting and closing, etc.).
Importantly, most utterances are emotionally neutral.
As the classifier has no `neutral' label, it assigns spurious emotions to these neutral utterances.

\subsection{Relationship between Rating and User Emotion}
\label{sec:analysis_emotion}

To understand users' emotions and how they relate to our bot's performance, we replicated our experiment for dialogue act labels by applying a regression analysis, to the emotion classifier labels and the ultimate rating of each conversation. 

Before performing this analysis, we removed all one-word utterances, since we assumed that these would not contain any emotion, and 66 common utterances that accounted for 40\% of responses (e.g. \textit{yes} and \textit{no}), assuming that they were also neutral. 

Figure \ref{fig:emotion_coef} shows that, as we would expect, positive emotions have the largest positive coefficients and negative emotions have the largest negative ones. 
A possible explanation for the anomalies (e.g. "terrified" having a relatively large positive coefficient) is that the emotion classifier strongly associates certain entities with emotions and struggles to recognize when these entities are used in different contexts. 
For example, it associates "tiger" with "terrified", even when "tiger" is in a positive context such as "I like tigers."

\section{Offensive User Experiment Details}
\label{app:offensive}
\subsection{Offense Type Detection}
\label{sec:appendix-offensive-user-offense-detection}
To determine the offense type, we hand-labeled 500 most common offensive utterances, which accounted for 53\% of all the offensive utterances we collected to the date. We used 6 categories: sexual, insult, criticism, inappropriate topic, bodily harm and error. To classify the user utterance into one of these categories, we built regular expressions checking if the given user utterance contains one of the hand-labeled examples for an offense type. We then used the offense type to contextualize our \texttt{COUNTER+PROMPT} and \texttt{EMPATHETIC+PROMPT} responses. 

\subsection{Response Strategy Configurations}
\label{sec:appendix-offensive-user-configurations}
This section gives a detailed description of the configurations used in the Offensive User experiments (\secref{sec:offensive_user_rg}).

\begin{enumerate}
    \item \texttt{WHY}: We ask the user why they made the offensive utterance (and this forms the entire bot utterance for the turn).
    The Offensive User RG responds with \textit{OK} to whatever the user says next, then hands over to another RG to supply a prompt.
    For example: \textbf{Bot}: \textit{Why did you say that?}, \textbf{User}: \textit{because you weren't understanding me}, \textbf{Bot}: \textit{OK. So, who's your favorite musician?}
    
    \item \texttt{WHY+NAME}: Same as \texttt{WHY}, but we append the user's name to the end of the bot utterance. 
    For example: \textit{Why did you say that, Peter?}

    \item \texttt{AVOIDANCE}: The bot politely avoids talking about the offensive topic, e.g. \textit{I'd rather not talk about that.} 
    This forms the entire utterance for the turn; the bot does not give any prompt to steer the conversation in a different direction.
    
    \item \texttt{AVOIDANCE+NAME}: Same as \texttt{AVOIDANCE}, but we append the user's name to the bot utterance. For example: \textit{I'd rather not talk about that\ Peter.}
    
    \item \texttt{AVOIDANCE+PROMPT}: Same as \texttt{AVOIDANCE}, but we also give a prompt to change the topic.
    For example: \textit{I'd rather not talk about that. So, who's your favorite musician?}
    
    \item \texttt{AVOIDANCE+NAME+PROMPT}: Same as \texttt{AVOIDANCE+NAME}, but append a prompt to the end of the utterance. 
    For example: \textit{I'd rather not talk about that, Peter. So, who's your favorite musician?}
    
    \item \texttt{COUNTER+PROMPT}: Strategy suggested by \citet{brahnam2005strategies} and evaluated by \citet{chin2020empathy}. In our interpretation of the strategy, we point out the inappropriate nature of the user utterance to the user, and attempt to move on to a different topic. For example, \textit{That is a very suggestive thing to say. I don't think we should be talking about that. Let's move on. So, who's your favorite musician?}
    
    \item \texttt{EMPATHETIC+PROMPT}: Strategy suggested by \citet{chin2020empathy} as the most appropriate strategy. In our interpretation of the strategy, we empathize with the user's desire to talk about inappropriate topics, and attempt to move on to a different topic. For example, \textit{If I could talk about it I would, but I really couldn't. Sorry to disappoint. So, who's your favorite musician?}

\end{enumerate}
For \texttt{COUNTER+PROMPT} and \texttt{EMPATHETIC+PROMPT}, unlike \cite{brahnam2005strategies} and \cite{chin2020empathy}, we handwrote responses tailored for the type of the user offense (e.g. sexual, curse, etc.)  

Note that \texttt{WHY} and \texttt{AVOIDANCE} cannot be combined because they are contradictory, and \texttt{WHY} and \texttt{CHANGE} cannot be combined because we cannot ask two questions in one turn.

\section{Opinion Agreement Policy Details}
\label{sec:opinion_policies_details}

Here we will show demonstrations of the 3 different Opinion Policies. All the demonstrations below are generated by researchers and are not actual user conversations. 

\subsection{ALWAYS\_AGREE Policy}
We begin by asking the user their opinion on the entity. Then, we agree with the user on their sentiment on the entity and share our reason, and ask the user their reason for liking/disliking the entity. After that, we propose another reason for liking/disliking the entity, and ask whether the user agrees with our reason or not. Finally, we switch to a related entity and repeat the same steps as described before. A demonstration of the policy is shown below.
\begin{quote}
    \textbf{User}: \textit{let's talk about cats} \\
    \textbf{Bot}: \textit{Ok! Do you like cats?} \\
    \textbf{User}: \textit{cats are my favorite} \\
    \textbf{Bot}: \textit{Sounds like you like cats. Me too! I feel like cats are so fluffy. What about you?} \\
    \textbf{User}: \textit{i like cats because they are very independent} \\
    \textbf{Bot}: \textit{That's so true. That reminds me of another reason I love cats. I feel like they treat you as an equal. Do you agree?} \\
    \textbf{User}: \textit{i totally agree}\\
    \textbf{Bot}: \textit{What about dogs? Do you like dogs?}\\
    \textbf{User}: \textit{yes}\\
    \textbf{Bot}: \textit{Me too! You know, I think the reason I'm a fan of dogs is because they love you back no matter what. What do you think?}\\
    \textbf{User}: \textit{yeah i feel the same way}\\
    \textbf{Bot}: \textit{Totally. I also like dogs because of how nifty they are. Do you feel the same way?}\\
    \textbf{User}: \textit{Yeah you are right they are pretty nifty.}\\
    \textbf{Bot}: \textit{Thanks for sharing! It's nice to know your likes and dislikes. Do you want to know more about dogs?}
\end{quote}

\subsection{LISTEN\_FIRST\_DISAGREE Policy}
We initially ask the user for their opinion and reason (i.e. "What do you think about cats?" followed by "Why do you like cats?"), then we disagree with the user by giving our opinion and reason. After that, we switch to a related entity and agree with the user on their sentiment on the new entity. A demonstration of the policy is shown below.
\begin{quote}
    \textbf{User}: \textit{let's talk about cats} \\
    \textbf{Bot}: \textit{Ok! Do you like cats?} \\
    \textbf{User}: \textit{cats are my favorite} \\
    \textbf{Bot}: \textit{What's your favorite thing about cats?} \\
    \textbf{User}: \textit{i like cats because they are very independent} \\
    \textbf{Bot}: \textit{That make sense. I have to be honest though, I'm not a big fan of cats actually. I feel like they don't respect my personal space. Can we agree on that?} \\
    \textbf{User}: \textit{yeah i guess you are right on that one}\\
    \textbf{Bot}: \textit{What about dogs? Do you like dogs?}\\
    \textbf{User}: \textit{yes}\\
    \textbf{Bot}: \textit{Me too! You know, I think the reason I'm a fan of dogs is because they love you back no matter what. What do you think?}\\
    \textbf{User}: \textit{yeah i feel the same way}\\
    \textbf{Bot}: \textit{Totally. I also like dogs because of how nifty they are. Do you feel the same way?}\\
    \textbf{User}: \textit{Yeah you are right they are pretty nifty.}\\
    \textbf{Bot}: \textit{Thanks for sharing! It's nice to know your likes and dislikes. Do you want to know more about dogs?}
\end{quote}

\subsection{CONVINCED\_AGREE Policy}
This is a stronger version of the previous policy in which we immediately disagree with the user and give our opinion and reason (i.e. "What do you think about cats?" followed by "That's good to know, but I don't quite get the appeal of cats. I feel like they are too independent. But I want to hear why you like cats"). After receiving an answer from the user, we agree with the user and appear convinced. After that, we switch to a related entity and agree with the user on their sentiment on the new entity. A demonstration of the policy is shown below.
\begin{quote}
    \textbf{User}: \textit{let's talk about cats} \\
    \textbf{Bot}: \textit{Ok! Do you like cats?} \\
    \textbf{User}: \textit{cats are my favorite} \\
    \textbf{Bot}: \textit{Glad to meet a fan of cats! I have to be honest though, I'm not a big fan of cats actually. I feel like they don't respect my personal space. But I'm interested to hear why you like cats?} \\
    \textbf{User}: \textit{i don't agree i like that about cats because that's how they show their love} \\
    \textbf{Bot}: \textit{That make sense. Now that I think about it, there are a few things I like about cats. For example, they are very independent. What do you think?} \\
    \textbf{User}: \textit{yeah that's right.}\\
    \textbf{Bot}: \textit{What about dogs? Do you like dogs?}\\
    \textbf{User}: \textit{yes}\\
    \textbf{Bot}: \textit{Me too! You know, I think the reason I'm a fan of dogs is because they love you back no matter what. What do you think?}\\
    \textbf{User}: \textit{yeah i feel the same way}\\
    \textbf{Bot}: \textit{Totally. I also like dogs because of how nifty they are. Do you feel the same way?}\\
    \textbf{User}: \textit{Yeah you are right they are pretty nifty.}\\
    \textbf{Bot}: \textit{Thanks for sharing! It's nice to know your likes and dislikes. Do you want to know more about dogs?}
\end{quote}

\end{document}